\documentclass{article}

 \usepackage[main, final]{neurips_2026}
\usepackage[utf8]{inputenc} % allow utf-8 input
\usepackage[T1]{fontenc}    % use 8-bit T1 fonts
\usepackage{hyperref}       % hyperlinks
\usepackage{url}            % simple URL typesetting
\usepackage{booktabs}       % professional-quality tables
\usepackage{amsfonts}       % blackboard math symbols
\usepackage{nicefrac}       % compact symbols for 1/2, etc.
\usepackage{microtype}      % microtypography
\usepackage[table]{xcolor}         % colors
\usepackage{natbib}
\usepackage{graphicx}
\usepackage{amsmath}
\usepackage{amssymb}
\usepackage{wrapfig}
\usepackage{pifont}
\usepackage{makecell}
\usepackage{multirow}
\usepackage{geometry}
\usepackage{caption}
\usepackage{subcaption}
\usepackage[most]{tcolorbox}
\usepackage{tikz}
\usetikzlibrary{tikzmark, decorations.pathreplacing, calc}
\newcommand{\LineComment}[1]{{\color{blue}\textit{$\triangleright$ #1}}}

\newcommand{\xmark}{\ding{55}} 
\newcommand{\cmark}{\ding{51}}

\definecolor{GrayRow}{HTML}{EEF0F4}
\definecolor{GreenRow}{HTML}{F0F3EA}
\definecolor{citeblue}{rgb}{0.368,0.507,0.71}
\definecolor{refgreen}{RGB}{67, 125, 122}
\definecolor{urlblue}{HTML}{47b8d1}
\definecolor{algcomment}{RGB}{34, 139, 34}
\definecolor{mycommentcolor}{RGB}{69, 176, 176}
\renewcommand{\LineComment}[1]{\textcolor{mycommentcolor}{$\triangleright$ #1}}

\newcommand{\drop}[1]{\rlap{$^{\textcolor{red!70}{\downarrow \textbf{#1}}}$}}
\newcommand{\up}[1]{\rlap{$^{\textcolor{green!70!black}{\uparrow \textbf{#1}}}$}}

\hypersetup{
    final,
    colorlinks,
    linkcolor=refgreen,
    citecolor=citeblue,
    urlcolor=urlblue,
}

\definecolor{promptbackground}{HTML}{E6E6E6}
\definecolor{prompttitlebg}{HTML}{F0F0F0}

\newtcolorbox{promptbox}[1]{
    enhanced,                 % Use the enhanced skin for a clean look
    title=#1,                 % The title is taken from the argument
    colback=prompttitlebg,            % The main content background is white
    colframe=black,     % Use our custom gray for the frame
    colbacktitle=promptbackground, % Use our custom light gray for the title background
    coltitle=black,           % The title text is black
    fonttitle=\bfseries,      % Make the title font bold
    boxrule=1pt,              % Set the border thickness to 1 point
    arc=2mm,                  % Create rounded corners with a 2mm radius
}

\title{SEED: Self-Speculative Decoding via Implicit Encoder–Decoder}

\author{%
  Hankun Lin\thanks{Equal contribution.}, \ Patrick Pynadath\footnotemark[1], \ Ruqi Zhang \\
  Department of Computer Science, Purdue University, USA \\
  \texttt{\{lhankun, ppynadat, ruqiz\}@purdue.edu} \\
}

\begin{document}

\maketitle

\newcommand{\methodname}[0]{SEED}

\begin{abstract}
Self-speculative decoding accelerates large language model (LLM) inference by drafting tokens from the target model itself, but faces a sharp tradeoff between the quality and cost of the draft. Early-exit methods produce drafts cheaply by terminating computation at intermediate layers, but forgo the deeper representations that later layers provide and thus suffer in draft quality. Multi-token prediction preserves draft quality by emitting from the model's final hidden states, but pays for a full forward pass to produce those states at every drafting step.
We propose \emph{\textbf{s}elf-sp\textbf{e}culative \textbf{e}ncoder-\textbf{d}ecoder} (\textbf{\methodname}), a self-speculative method that obtains high-quality drafts cheaply by reusing the deep contextual representations already computed during verification. We reinterpret the standard decoder-only transformer as an implicit encoder–decoder: the first layers (encoder) build deep contextual representations, and the last few layers (decoder) emit tokens from them. Encoding and verification are merged into a single step: verification is performed by the full encoder–decoder, and the contextual representations of the verified prefix are cached for reuse during drafting. Drafting is therefore very fast: between verifications, the lightweight decoder drafts multiple tokens autoregressively, each conditioned on the cached representations and on preceding drafts.
Experiments across multiple benchmarks show that \methodname\ achieves up to 2.7$\times$ average speedup on 4B-scale models, outperforming both early-exit and MTP-style self-speculative baselines and running 28\% faster than the state-of-the-art EAGLE-3, while preserving or even improving the generation quality of standard autoregressive fine-tuning. Code is available at \url{https://github.com/lhk2004/SEED}.
\end{abstract}

\section{Introduction}
\begin{figure}[t]
\begin{center}
\includegraphics[width=\textwidth]{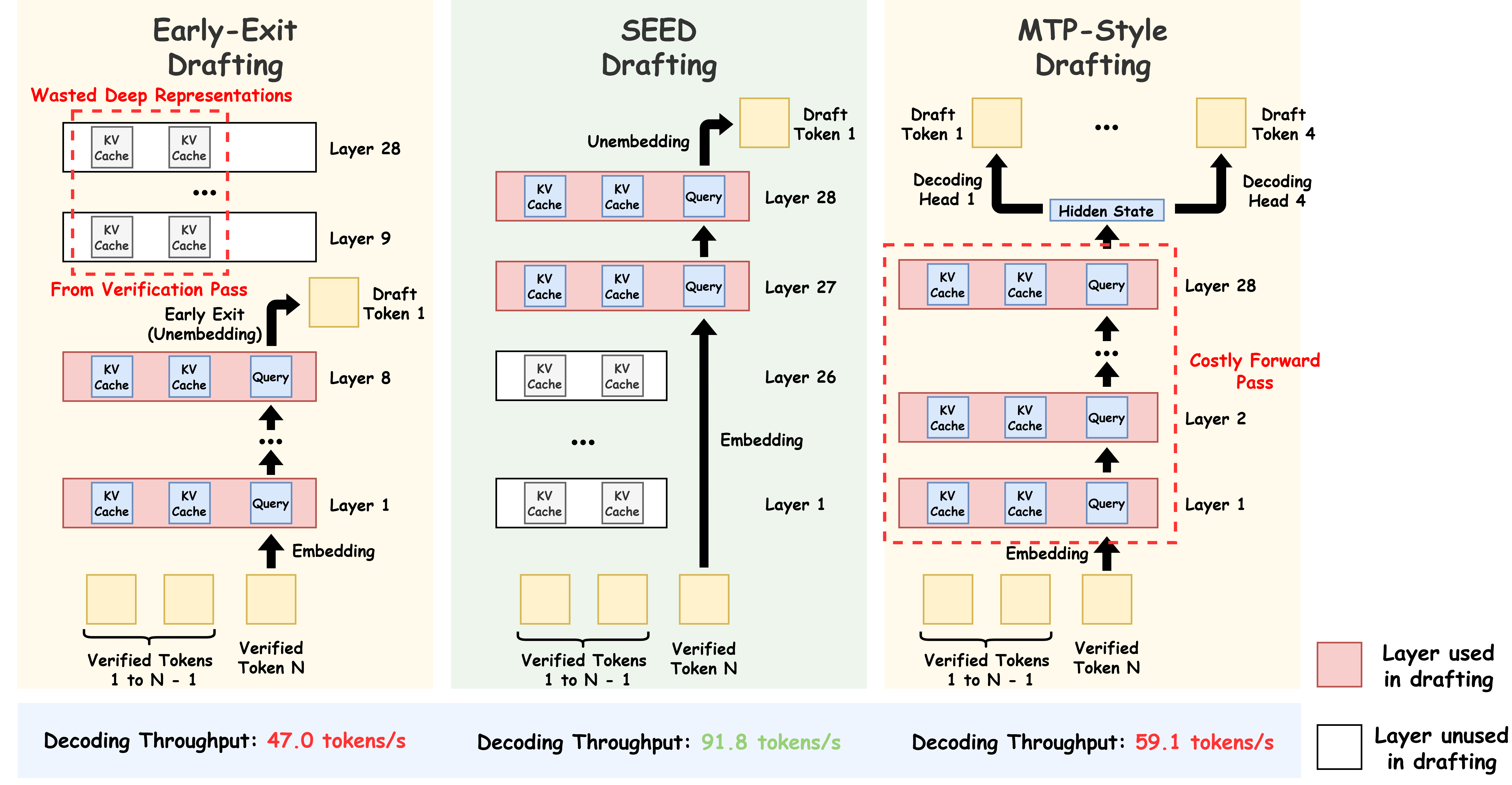}
\end{center}
\caption{\textbf{Overview of \methodname.} \methodname\ partitions a $28$-layer model into a $26$-layer encoder and a $2$-layer decoder. During drafting, the decoder generates drafts by cross-attending to the deep contextual KV cache produced during the previous verification step, avoiding repeatedly invoking the costly encoder. This design contrasts with two existing self-speculative decoding paradigms: early-exit drafting (instantiated using LayerSkip~\citep{elhoushi2024layerskip}) discards the deep representations from later layers, sacrificing draft quality, while multi-token prediction (instantiated using Apple MTP~\citep{samragh2025your}) preserves draft quality by maintaining full representational depth but still requires a full forward pass at every drafting step. On GSM8K with draft length $4$, \methodname\ achieves substantially higher decoding throughput than early-exit drafting ($\sim2.0\times$) and MTP ($\sim1.6\times$).}
\label{fig:main}
\end{figure}

Large language model (LLM) inference is memory-bound rather than compute-bound~\citep{pope2023efficiently, fu2024break}: generating each token requires a full forward pass through the model, yet that pass produces only a single output, leaving the GPU's parallel compute capacity underutilized. Speculative decoding~\citep{leviathan2023fast, chen2023accelerating} amortizes this cost by using a cheap draft model to propose multiple candidate tokens that are verified in parallel by the target model in a single forward pass. However, maintaining a separate, well-aligned drafter complicates deployment.
This has motivated self-speculative decoding, where drafting and verification are handled within a single unified model.

Self-speculative methods face a sharp tradeoff between the quality and cost of the draft.
Early-exit or layer-skipping methods~\citep{zhang2024draft, elhoushi2024layerskip, xia2025swift, chen2025clasp} terminate the forward pass at an intermediate layer, making drafting cheap but forcing the drafter to predict from shallow representations, which limits draft quality.
By contrast, multi-token prediction (MTP) methods~\citep{gloeckle2024better, samragh2025your} emit drafts from the model's final hidden states, preserving representational depth but requiring a full forward pass at every drafting step.
This raises a question: \textit{can drafts be as deep as the full model yet as cheap as a few layers?}

Our answer adapts the encoder-decoder amortization principle developed in diffusion language models~\citep{arriola2025encoder} to autoregressive speculative decoding. We view the standard decoder-only transformer as an implicit encoder-decoder: the first layers act as an encoder that builds deep contextual representations of the prefix, while the last few layers act as a decoder that autoregressively emits the next token from those representations.
The crucial observation is that under this view, verification already does all the encoding work needed for subsequent drafting --- the deep representations the drafter needs are already in the KV cache. Building on this, we propose \emph{\textbf{s}elf-sp\textbf{e}culative \textbf{e}ncoder-\textbf{d}ecoder} (\textbf{\methodname}), a self-speculative decoding method that trains the lightweight decoder to generate new draft tokens by conditioning on the verifier's deepest features, without invoking the encoder again until the next verification step. Each draft token therefore costs only a thin-decoder forward pass yet conditions on the verifier's full-depth representation of the prefix. Unlike prior methods that require careful architectural modifications or nuanced training procedures, \methodname\ is remarkably simple: it adds only a single auxiliary loss on top of standard supervised fine-tuning. Figure~\ref{fig:main} provides an overview of \methodname.

Across math, coding, general knowledge, and summarization benchmarks, \methodname\ achieves up to a \textbf{2.7$\times$ average speedup on 4B-scale models} while matching or improving the generation quality of standard autoregressive fine-tuning. It consistently outperforms both early-exit and MTP-based self-speculative baselines and runs \textbf{28\% faster than the state-of-the-art EAGLE-3}. Moreover, our analysis reveals that, like MTP methods, \methodname\ encourages the model to encode more predictive information about future tokens, which contributes to improving both draft quality and downstream task performance.

\section{Related Work}
There has been substantial interest in accelerating large language model (LLM) inference. While classical speculative decoding requires a small auxiliary model or multiple heads \citep{stern2018blockwise, cai2024medusa, anknerhydra, li2025eagle3}, most relevant to our work are self-speculative decoding methods, which avoid using an additional draft model by drafting from the target model itself. These methods fall broadly into two categories, multi-token prediction (MTP) and exiting early by skipping layers.

\paragraph{Multi-Token Prediction.}
Multi-token prediction (MTP) accelerates inference by predicting multiple tokens in a single forward pass, increasing the model's output per pass while keeping per-pass cost roughly fixed. Originally motivated by denser training signals, early work modified the pretraining objective to enable parallel token prediction~\citep{qi2020prophetnet, gloeckle2024better, liu2024deepseek}, but reliance on pretraining from scratch makes these approaches impractical for accelerating deployed models. More recent work adapts the MTP pipeline for faster inference on existing models~\citep{bhendawade2024speculative, samragh2025your, mehra2025multi, liumtp, cai2025fastmtp, draxler2025parallel, kirchenbauer2026multi, zhao2026self}. A separate line replaces the standard MTP formulation with alternative decoding schemes such as Jacobi decoding~\citep{kou2024cllms, hu2025fast} or diffusion-based generation~\citep{arriola2025encoder, liu2025tidar, chen2026dflash}, but these either sacrifice quality by breaking causal dependencies or substantially complicate the training objective.

\paragraph{Early-Exit Methods.}
Early-exit methods reuse components of the verifier itself as the drafter, typically skipping later layers and using only the initial layers for drafting. This can be achieved through additional training~\citep{elhoushi2024layerskip, liu2024kangaroo} or as a purely inference-time decision~\citep{zhang2024draft, xia2025swift, chen2025clasp, zarch2025context}. \methodname\ shares the layer-skipping spirit but inverts the design: rather than skipping later layers, we skip the initial layers and draft from only the final layers of the base model. This asymmetric choice lets the drafter condition on full-depth contextual representations, producing higher-quality draft tokens than early-exit alternatives.

For a more detailed comparison with existing LLM acceleration methods, see Appendix~\ref{sec:add_related_works}.

\section{Preliminaries}
We represent a standard transformer as a function $f$ that maps an input sequence $X = (X_1, X_2, \dots, X_n)$ of tokens from a vocabulary $\mathcal{V}$ to a probability distribution over the next token, $f(X) \in \triangle(\mathcal{V})$. The transformer is composed of $l$ layers $(f_1, f_2, \dots, f_l)$. Each layer $f_i$ takes as input the hidden states from the previous layer, $H^{i-1} = (H_1^{i-1}, H_2^{i-1}, \dots, H_n^{i-1})$, and outputs updated representations $H^i = (H_1^i, H_2^i, \dots, H_n^i)$. Each self-attention layer additionally computes keys and values that are cached to avoid redundant computation;
we denote the KV cache available before processing token $X_t$ at layer $i$
as $KV^i_{<t}$.
For brevity, we omit explicit notation for KV cache updates in the method description that follow.

The first layer $f_1$ operates on the token embeddings $H^0 = W_{\text{in}} X$, where $W_{\text{in}}$ is the embedding matrix. The output of the final layer, $H^l$, is projected back to vocabulary space via a second embedding matrix $W_{\text{out}}$ to obtain the next-token distribution,
\begin{align}
P(\cdot \mid X_{\le n}) = \mathrm{softmax}(H_n^l W_{\text{out}}^\top).
\end{align}

Autoregressive transformer language models are trained by minimizing the standard next-token cross-entropy loss
\begin{align}
\mathcal{L}_{\text{CE}} = \mathbb{E}_{X \sim \mathcal{D}} \Bigg[ \sum_{j=1}^{|X|} -\log f(X_j \mid X_{<j}) \Bigg],
\end{align}
where $\mathcal{D}$ denotes the training data distribution.

\section{Self-Speculative Encoder-Decoder}
\label{sec:methodology}
We propose \textbf{s}elf-sp\textbf{e}culative \textbf{e}ncoder-\textbf{d}ecoder (\textbf{\methodname}), a self-speculative decoding method
that obtains high-quality drafts cheaply while keeping the model architecture unchanged. We first introduce the encoder-decoder reinterpretation of the standard decoder-only transformer, and explain why it enables cheap yet accurate drafting. We then present the training algorithm, which augments the standard next-token prediction objective with an additional speculative drafting objective. This trains the decoder to draft from raw token embeddings while cross-attending to cached deep representations. Finally, we describe the inference procedure, including a fixed-length drafting and a dynamic drafting with tree-based verification.

\subsection{Method}
A key observation underlying \methodname\ is that a standard decoder-only transformer can be naturally interpreted as an implicit encoder-decoder architecture. This reinterpretation exposes a structural asymmetry that we exploit to make cheap yet accurate drafting. 

\newcommand{\encoder}[0]{f_{enc}}
\newcommand{\decoder}[0]{f_{dec}}
\newcommand{\kvcache}[2]{KV_{#1}^{#2}}
\newcommand{\hiddenreps}[2]{H_{#1}^{#2}}

\paragraph{Decoder-Only Transformers as Implicit Encoder-Decoders.}
Consider a transformer with $l$ layers $(f_1, \dots, f_l)$. We partition the model at some layer $l'$, treating the first $l'$ layers as an \emph{encoder} that builds deep contextual representations of the prefix, and the remaining layers as a \emph{decoder} that emits the next token from those representations. The encoder typically accounts for the majority of layers (e.g., $l'=26$ out of $l=28$ in our experiments).

Formally, we define $\encoder = f_{1:l'}$ and $\decoder = f_{l'+1:l}$, with $KV^{enc}_{<n}$ and $KV^{dec}_{<n}$ denoting the cached keys and values
from all positions preceding $n$ in the encoder and decoder respectively.
After processing token $X_n$, the corresponding caches are updated to
$KV^{enc}_{\le n}$ and $KV^{dec}_{\le n}$.
For a context sequence $X_{\le n}$, the encoder produces a deep contextual embedding of the current input token $X_n$:
\begin{align}
H_n^{l'} = \encoder(\cdot \mid X_n, \kvcache{<n}{enc}),
\end{align}
and the decoder predicts the next token from this representation while attending to the cached context:
\begin{align}
X_{n+1} \sim
\mathrm{softmax}\!\left(
\decoder(\cdot \mid H_n^{l'}, \kvcache{<n}{dec})
\right).
\end{align}
This decomposition is purely conceptual as it is exactly equivalent to standard autoregressive decoding, but it makes a structural asymmetry explicit: the encoder accounts for most per-token computation while the decoder is a thin stack of only a few layers.

Standard autoregressive generation does not exploit this asymmetry, since the encoder is re-invoked for every new token. We propose to exploit it for self-speculative decoding: because context evolves slowly, encoding can be performed infrequently, and the decoder alone can emit tokens conditioned on a fixed contextual representation.

\paragraph{Verification Already Does the Encoder's Work for Drafting.}
Remarkably, we do not need to run the encoder separately to obtain the contextual representation. The verification step already does this work.

To verify a batch of proposed drafts, the full encoder-decoder runs on these drafts and produces $\kvcache{<n}{dec}$, the deep contextual representations up to the last verified draft token $X_{n-1}$. Because the verifier has processed the entire prefix $X_{<n}$ during this forward pass, it should be able to predict another bonus token $X_n$ through the same pass. Thus, the cache contains deep contextual representations only for tokens preceding $X_n$, which are precisely the representations needed for subsequent drafting.
Verification and encoding therefore merge into a single step: one forward pass of the full model simultaneously verifies the drafts and prepares the deep contextual cache that the drafter consumes in the next round.

\paragraph{Cheap Drafting Between Verifications.}
With the deep representations already cached during verification, the decoder no longer requires a fresh encoder call to begin drafting. We instead feed the raw token embedding $\hiddenreps{n}{0} = W_{in} X_n$ of the latest token directly into the decoder, which conditions on the cached deep representations through cross-attention to predict the next token
\begin{align}
X_{n+1} \sim 
\mathrm{softmax}\!\left(
\decoder(\cdot \mid \hiddenreps{n}{0}, \kvcache{< n}{dec})
\right).
\end{align}
Between verifications, drafting requires only the lightweight decoder. The encoder is invoked again only when the next batch of drafts is verified, at which point it produces both verification and the refreshed cache.

Note that we use the term \emph{cross-attention} as shorthand for attention whose queries come from raw token embeddings and whose keys and values come from the encoder-produced cache. This is implemented as standard self-attention over the KV cache and does not denote a separate encoder–decoder cross-attention module. We adopt the term "cross-attention" only to emphasize that in decoder, the cache carries full-depth, encoder-processed representations while the incoming query carries only a shallow raw embedding, playing a role analogous to conventional encoder–decoder cross-attention where decoder queries attend to encoder-derived keys and values.

\subsection{Training Algorithm}
We now describe how to train the decoder to draft accurately from raw embeddings while cross-attending to cached deep representations. We focus on the typical scenario in which \methodname\ is integrated into a standard supervised fine-tuning pipeline: starting from a pre-trained base LLM, we fine-tune it on task data to improve task performance while simultaneously enabling self-speculative decoding. The same procedure also applies when starting from an already fine-tuned model and continuing training purely to enable inference acceleration, which we explore in Appendix~\ref{sec:retrofit}.

\paragraph{Standard Autoregressive Objective.}
To preserve the model's generation quality, we retain the standard next-token prediction objective, re-expressed under our encoder-decoder decomposition:

\begin{align}
\mathcal{L}_{CE}
=
\mathbb{E}_{X \sim \mathcal{D}}
\left[
\sum_{j=1}^{|X|}
-
\log
\decoder
\left(
X_j
\mid
H_{j-1}^{l'},
\kvcache{<(j-1)}{dec}
\right)
\right].
\end{align}

This loss exactly recovers standard autoregressive training, ensuring that the full encoder-decoder (the verifier) behaves identically to a standard autoregressive model.

\paragraph{Speculative Drafting Objective.}
At inference, the drafter must predict the next token conditioned on (i) deep contextual representations $\kvcache{<n'}{dec}$ produced by the verifier for tokens up to the last verified position $n'$, and (ii) only the raw token embeddings $\hiddenreps{n' \le \cdot \le (j-1)}{0} = W_{in} X_{n' \le \cdot \le (j-1)}$ for the last verified context token and any subsequent draft tokens, since invoking the encoder on draft tokens would defeat the purpose of cheap drafting. To train the decoder to predict the next token given this input, we introduce a speculative drafting objective:

\begin{align}
\label{eq:sepc_obj}
\mathcal{L}_{spec}
=
\mathbb{E}_{X \sim \mathcal{D}}
\left[
\sum_{j=1}^{|X|}
% \sum_{n'=1}^{j-1}
-
\log
\decoder
\left(
X_j
\mid
\hiddenreps{n' \le \cdot \le (j-1)}{0},
\kvcache{<n'}{dec}
\right)
\right].
\end{align}

\paragraph{Training with Block Attention.} Optimizing $\mathcal{L}_{spec}$ requires choosing a boundary $n'$ for each target token $X_j$ that splits the conditioning context into cached deep representations $KV^{dec}_{<n'}$ and raw token embeddings $H^0_{n' \le \cdot \le (j-1)}$. This boundary directly mirrors the inference-time situation we want the decoder to handle: between two verification steps, the decoder must draft several new tokens on its own using only their raw embeddings, while cross-attending to the deep KV cache produced during the previous verification, and only after the next verification do these draft positions get committed to deep contextual representations. To expose the decoder to exactly this conditioning pattern during training, we partition each training sequence into contiguous, non-overlapping blocks of $b$ tokens, where each block plays the role of a sequence of drafts that the decoder must produce alone between two verifications. For every target token $X_j$, we set $n'$ to the index of the first token in the block containing $X_{j-1}$ (i.e., $n' = \lfloor (j-2)/b \rfloor \times b + 1$). This means that within each block, the decoder sees only raw token embeddings for the current block's tokens (via causal self-attention) and cross-attends to the deep KV cache $KV^{dec}_{<n'}$ produced by all preceding blocks. Because $n'$ is the same for every token within a block, all blocks can be processed in parallel.

\begin{figure}[t]
\begin{center}
\includegraphics[width=\textwidth]{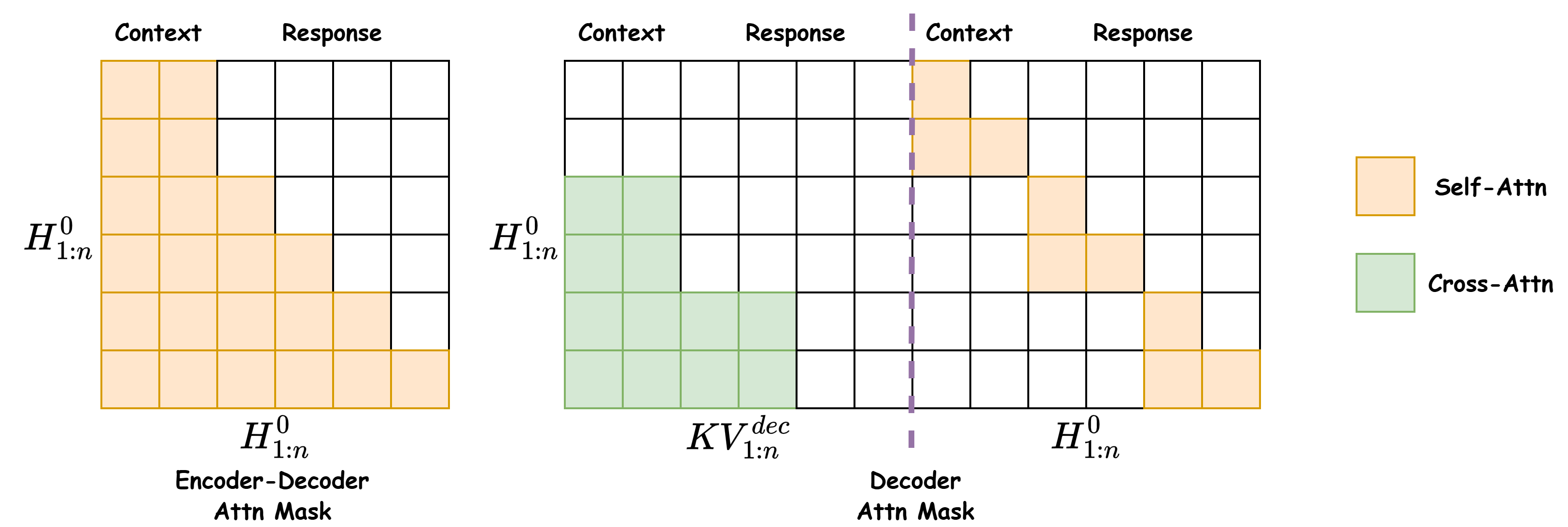}
\end{center}
\caption{Example training-time attention masks for $n = 6$ tokens with block size $b = 2$. \textbf{Left:} In the joint encoder-decoder verifier, context tokens use bidirectional attention for richer encoding, while response tokens use causal attention for autoregressive drafting. \textbf{Right:} In the decoder, input token embeddings ($H^0_{1:n}$) attend causally to tokens within the same block via self-attention, and additionally cross-attend to the decoder's own KV cache ($KV^{dec}_{1:n}$) of all preceding blocks.}
\label{fig:train-attn-mask}
\end{figure}

Concretely, training proceeds in two stages within each gradient step. First, the full encoder–decoder processes the entire sequence in a single forward pass, computing both the standard autoregressive loss $\mathcal{L}_{CE}$ and caching the decoder's KV representations $KV^{dec}_{1:n}$ at every position. Second, the decoder alone processes all blocks in parallel using the block-attention mask shown in Figure~\ref{fig:train-attn-mask}: within each block, tokens attend causally to other tokens in the same block via self-attention over raw embeddings $H^0$, and additionally cross-attend to $KV^{dec}$ entries from all preceding blocks (but not the current block). We adapt this block-wise training strategy from~\cite{arriola2025encoder}, but use \emph{block-causal} masks throughout rather than the block-bidirectional masks: response tokens in the encoder-decoder attend causally, preserving the verifier as a valid autoregressive model, and tokens within each decoder block attend causally so the drafter can generate tokens autoregressively. This preserves the autoregressive structure on both sides of the encoder-decoder split, which their block-diffusion formulation does not require.

\paragraph{Final Training Objective.}
Our final training objective combines the standard autoregressive loss with the speculative drafting loss through a weighted average:

\begin{align}
\label{eq:final_obj}
\mathcal{L}
=
\frac{\mathcal{L}_{CE} + \lambda \mathcal{L}_{spec}}{1 + \lambda},
\end{align}

where $\lambda$ controls the relative weight of the speculative drafting objective. In our experiments, we set $\lambda = 1.0$. Section~\ref{sec:ablation} studies the impact of varying $\lambda$, and the complete training procedure is summarized in Algorithm~\ref{alg:dual-decode-train-alg}.

\subsection{Inference Algorithm}
\textbf{Fixed-Length Drafting.\space}
At inference time, the trained decoder $\decoder$ acts as a lightweight draft model, while the full encoder–decoder $\decoder \circ \encoder$ serves as the verifier. Decoding alternates between drafting and verification: starting from a verified prefix, $\decoder$ autoregressively generates $d$ draft tokens by cross-attending to the deep contextual representations ($KV^{dec}$) cached from the previous verification step, after which the full encoder-decoder verifies these tokens in a single forward pass. We summarize the complete inference procedure in Algorithm~\ref{alg:inference-alg}.

\textbf{Adaptive Drafting with Tree-Based Verification.\space}
Fixed-length drafting is sufficient to realize the core benefit of \methodname, but it leaves additional speedup on the table because the decoder's confidence varies across generation steps: drafting too few tokens under-utilizes a confident decoder, while drafting too many wastes computation when the decoder is uncertain. To better exploit cheap drafting, we adopt two complementary techniques from prior work. First, following~\citep{liu2024kangaroo, xia2025swift, zarch2025context}, we replace the fixed draft length with a confidence-based stopping rule, allowing the decoder to draft more tokens when confident and fewer when uncertain. Second, following~\cite{xia2025swift} and~\cite{ning2025cas}, we expand the draft into multiple sibling candidates via a top-$p$ rule rather than committing to a single token. Across drafting steps, this yields a tree of candidate tokens that the full model verifies in one forward pass using a tree-structured attention mask preserving the correct causal dependencies. 

Both techniques are particularly well-suited to \methodname. Because the decoder shares parameters with the verifier, its confidence reliably reflects acceptance rate, making the stopping rule effective. And because each drafting step is cheap (only a lightweight-decoder forward pass), the overhead of producing extra candidates is small relative to the gain in accepted tokens per verification. We use this adaptive tree-based drafting strategy in the main experiments unless otherwise specified. We defer further illustration and implementation details to Appendix~\ref{sec:algorithms}, with ablations in Appendix~\ref{sec:further_ablations}.

\section{Experiments}
\label{sec:experiments}
\subsection{Experimental Setup}
\label{sec:exp_setup}

\begin{table*}[t]
\centering
\caption{\textbf{Performance comparison on \texttt{Qwen3-1.7B-Base} and \texttt{Qwen3-4B-Base}.} 0-shot pass@1 accuracy (Acc.) is reported in \%, and decoding throughput (Tput) in tokens/s. For CNN/Daily Mail, we report ROUGE-1 (R-1), ROUGE-2 (R-2), and ROUGE-L (R-L). For methods that keep the target model frozen, we report only their throughput. Given the size of its released drafter weights, PARD is evaluated only on \texttt{Qwen3-4B-Base}. The highest throughput and average speedup are shown in \textbf{bold}, and the second highest results are \underline{underlined}.}
\label{tab:main-results}
\resizebox{\textwidth}{!}{% Scales the table to fit the page width
\setlength{\tabcolsep}{4pt} % Slightly reduce padding between columns so it shrinks less overall
\begin{tabular}{l@{\hspace{0.5em}}l cc cc cc cccc c}
\toprule
\multicolumn{2}{@{}l}{\textbf{Method}} & \multicolumn{2}{c}{\textbf{GSM8K}} & \multicolumn{2}{c}{\textbf{KodCode}} & \multicolumn{2}{c}{\textbf{ScienceQA}} & \multicolumn{4}{c}{\textbf{CNN/Daily Mail}} & Avg. \\
\cmidrule(lr){3-4} \cmidrule(lr){5-6} \cmidrule(lr){7-8} \cmidrule(lr){9-12}
\multicolumn{2}{l}{} & \makebox[4.2em]{Acc.} & Tput & \makebox[4.0em]{Acc.} & Tput & \makebox[3.5em]{Acc.} & Tput & \makebox[3.5em]{R-1} & \makebox[3.5em]{R-2} & \makebox[3.5em]{R-L} & Tput & Speedup \\
\midrule
\multicolumn{13}{c}{\texttt{Qwen3-1.7B-Base}} \\
\midrule
\multicolumn{2}{@{}l}{AR} & 56.3 & 37.9 & 66.5 & 36.0 & 93.2 & 36.6 & 38.6 & 17.7 & 28.6 & 34.2 & 1.0 $\times$ \\
\midrule
\multicolumn{13}{c}{\textit{\textbf{Common Acceleration Methods}}} \\
\midrule
\multicolumn{2}{@{}l}{E2D2} & 26.6\drop{29.7} & 67.6 & 40.6\drop{25.9} & 65.1 & 88.7\drop{4.5} & 68.1 & 32.8\drop{5.8} & 12.1\drop{5.6} & 23.3\drop{5.3} & 56.8 & 1.8 $\times$ \\
\multicolumn{2}{@{}l}{Apple MTP} & - & 59.1 & - & 54.6 & - & 56.6 & - & - & - & 37.5 & 1.4 $\times$ \\
\multicolumn{2}{@{}l}{EAGLE-3} & - & \underline{79.4} & - & \underline{70.4} & - & \underline{75.4} & - & - & - & \underline{60.7} & \underline{2.0} $\times$ \\
\midrule
\multicolumn{13}{c}{\textit{\textbf{Early-Exit Self-Speculative Methods}}} \\
\midrule
\rowcolor{GrayRow}
\multicolumn{2}{@{}l}{LayerSkip} & 42.2\drop{14.1} & 47.0 & 57.5\drop{9.0} & 57.0 & 91.7\drop{1.5} & 60.0 & 37.8\drop{0.8} & 16.9\drop{0.8} & 27.6\drop{1.0} & 42.9 & 1.4 $\times$ \\
\rowcolor{GrayRow}
\multicolumn{2}{@{}l}{SWIFT} & - & 43.0 & - & 41.2 & - & 40.2 & - & - & - & 36.2 & 1.1 $\times$ \\
\rowcolor{GrayRow}
\multicolumn{2}{@{}l}{DEL} & - & 50.1 & - & 60.0 & - & 62.2 & - & - & - & 38.6 & 1.5 $\times$ \\
\midrule
\rowcolor{GreenRow}
\multicolumn{2}{@{}l}{\methodname~(Ours)} & 57.1\up{0.8} & \textbf{91.8} & 68.8\up{2.3} & \textbf{105.3} & 95.5\up{2.3} & \textbf{111.7} & 39.2\up{0.6} & 18.0\up{0.3} & 29.1\up{0.5} & \textbf{63.2} & \textbf{2.6}$\times$ \\

\midrule
\multicolumn{13}{c}{\texttt{Qwen3-4B-Base}} \\
\midrule
\multicolumn{2}{@{}l}{AR} & 75.1 & 28.3 & 76.2 & 27.6 & 95.4 & 26.0 & 39.0 & 18.2 & 29.1 & 24.0 & 1.0 $\times$ \\
\midrule
\multicolumn{13}{c}{\textit{\textbf{Common Acceleration Methods}}} \\
\midrule
\multicolumn{2}{@{}l}{E2D2} & 35.5\drop{39.6} & 55.8 & 48.0\drop{28.2} & 53.5 & 93.5\drop{1.9} & 54.6 & 32.7\drop{6.3} & 11.5\drop{6.7} & 23.5\drop{5.6} & 40.3 & 1.9 $\times$ \\
\multicolumn{2}{@{}l}{Apple MTP} & - & 45.1 & - & 49.7 & - & 41.3 & - & - & - & 29.4 & 1.6 $\times$ \\
\multicolumn{2}{@{}l}{EAGLE-3} & - & \underline{64.4} & - & 54.1 & - & 56.0 & - & - & - & \textbf{49.9} & \underline{2.1} $\times$ \\
\multicolumn{2}{@{}l}{PARD} & - & 57.7 & - & \underline{65.4} & - & \underline{62.7} & - & - & - & 41.0 & \underline{2.1} $\times$ \\
\midrule
\multicolumn{13}{c}{\textit{\textbf{Early-Exit Self-Speculative Methods}}} \\
\midrule
\rowcolor{GrayRow}
\multicolumn{2}{@{}l}{LayerSkip} & 62.2\drop{12.9} & 34.2 & 71.1\drop{5.1} & 43.7 & 92.4\drop{3.0} & 39.8 & 38.7\drop{0.3} & 17.6\drop{0.6} & 28.5\drop{0.6} & 28.7 & 1.4 $\times$ \\
\rowcolor{GrayRow}
\multicolumn{2}{@{}l}{SWIFT} & - & 31.7 & - & 31.3 & - & 25.8 & - & - & - & 26.1 & 1.1 $\times$ \\
\rowcolor{GrayRow}
\multicolumn{2}{@{}l}{DEL} & - & 34.8 & - & 41.0 & - & 39.4 & - & - & - & 22.1 & 1.3 $\times$ \\
\midrule
\rowcolor{GreenRow}
\multicolumn{2}{@{}l}{\methodname~(Ours)} & 76.0\up{0.9} & \textbf{74.9} & 81.0\up{4.8} & \textbf{80.5} & 96.8\up{1.4} & \textbf{89.3} & 39.8\up{0.8} & 18.4\up{0.2} & 29.3\up{0.2} & \underline{43.3} & \textbf{2.7}$\times$ \\

\bottomrule
\end{tabular}%
}
\end{table*}

\textbf{Datasets \& Metrics.\space}
We demonstrate the effectiveness of \methodname\ by training task-specific models across a diverse set of benchmarks, covering mathematical reasoning, code generation, general knowledge question answering, and summarization: GSM8K~\citep{cobbe2021training}, KodCode~\citep{xu2025kodcode}, ScienceQA~\citep{lu2022learn}, and CNN/Daily Mail~\citep{nallapati2016abstractive}. For efficiency, we report the decoding throughput achieved by each method on every dataset with a single 48 GB NVIDIA A6000 GPU. For task quality, we report 0-shot pass@1 accuracy on GSM8K, KodCode, and ScienceQA, and ROUGE scores for CNN/Daily Mail. More details regarding dataset construction can be found in Appendix~\ref{sec:datasets_details}.

\textbf{Baselines.\space}
We compare our method against the standard autoregressive (AR) decoding baseline with SFT. We further include three groups of accelerated decoding approaches:  
(1) \textbf{Early-exit self-speculative decoding} methods, including LayerSkip~\citep{elhoushi2024layerskip}, SWIFT~\citep{xia2025swift}, and DEL~\citep{zarch2025context};  
(2) \textbf{Speculative decoding} with a separate draft model, where we adopt the state-of-the-art EAGLE-3~\citep{li2025eagle3} and PARD~\citep{an2026pard} (4B target model only); (3) \textbf{Diffusion language models} and \textbf{multi-token prediction (MTP)} methods, including E2D2~\citep{arriola2025encoder}, which adopts an encoder–decoder division similar to ours to freeze context understanding but applies it within a blocked diffusion framework (Appendix~\ref{sec:e2d2_comp} further clarifies our novelty over E2D2), and Apple MTP~\citep{samragh2025your}, a strong MTP baseline that mitigates the loss of local dependency modeling in standard MTP. All methods use greedy decoding for generation.

\textbf{Models.\space}
For methods that require training (AR, E2D2, LayerSkip, Apple MTP, and our method), we fine-tune the pre-trained \texttt{Qwen3-1.7B-Base} and \texttt{Qwen3-4B-Base} models~\citep{yang2025qwen3} using the training split of each task dataset and evaluate on the corresponding test split. \methodname\ uses only the last 2 transformer layers of the model as the decoder and is trained with a block size of 10, i.e., each sequence is partitioned into 10-token blocks when computing the speculative drafting objective. We perform ablations on these choices in Section~\ref{sec:ablation} and provide setup details for other methods in Appendix~\ref{sec:baseline_details}.

\subsection{Results}

\begin{figure}[!b]
\centering

\begin{minipage}[t]{0.48\textwidth}
\centering
\vspace{0pt}
\includegraphics[width=\linewidth]{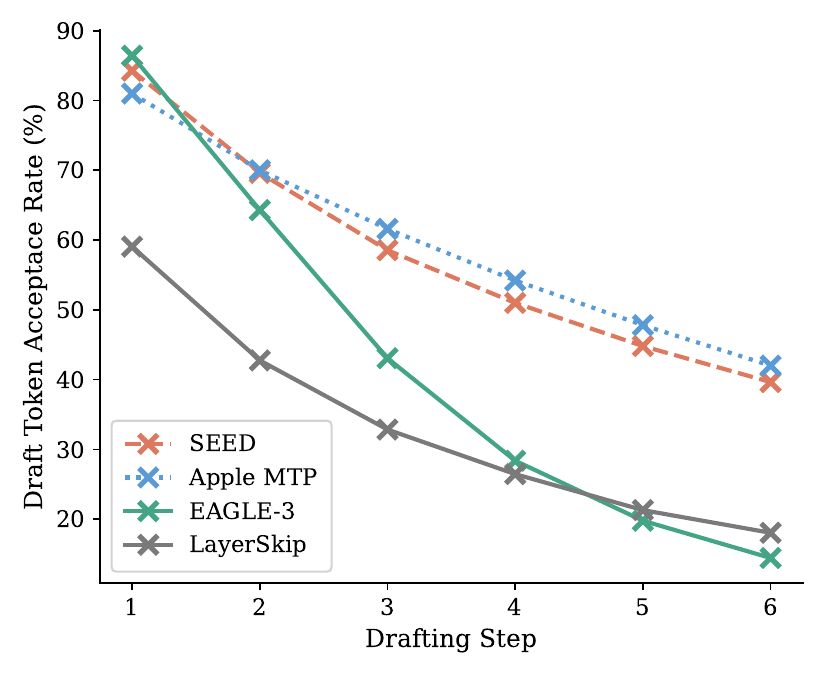}
\captionof{figure}{Draft token acceptance rates across drafting steps. For a fair comparison, all methods are evaluated under a fixed draft length of $6$, with only one candidate draft token kept at each step. Compared to LayerSkip and EAGLE-3, \methodname\ produces higher-quality drafts, both from deeper representations and from parameter sharing with the verifier. Compared to MTP, \methodname\ achieves comparable draft quality at much lower cost.}
\label{fig:draft_tok_acc_rate_per_pos}
\end{minipage}
\hfill
\begin{minipage}[t]{0.48\textwidth}
\centering
\vspace{0pt}
\includegraphics[width=\linewidth]{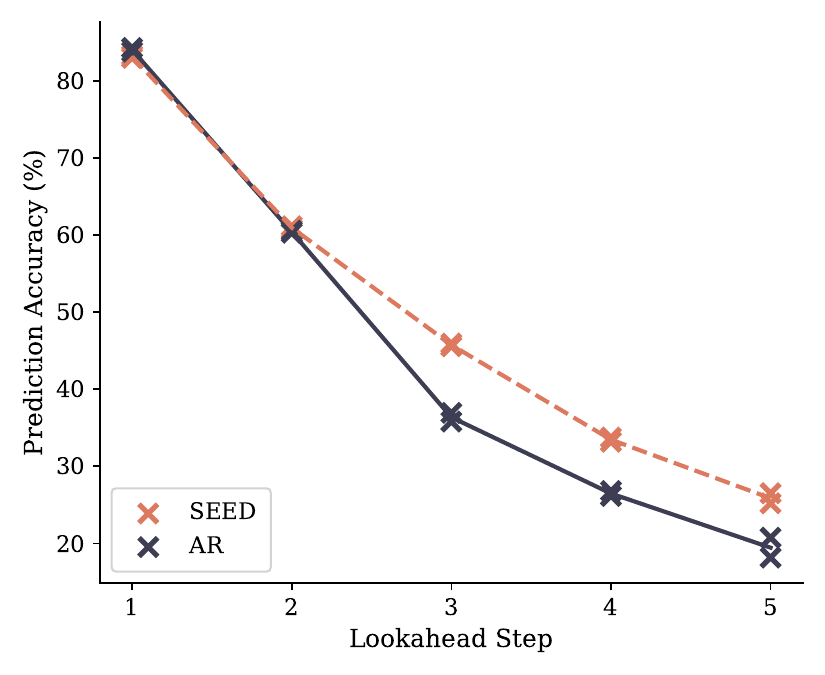}
\captionof{figure}{Prediction accuracy of trained linear probes in relation to lookahead step for two runs with different seeds. Compared to standard AR training, \methodname\ forces the encoder to produce representations that are more predictive of distant future tokens.}
\label{fig:enc_hid_pred}

\end{minipage}

\end{figure}

We present the main evaluation results in Table~\ref{tab:main-results}. Overall, \methodname\ delivers substantial inference acceleration. Across all benchmarks, \methodname\ achieves an average decoding speedup of \textbf{2.6$\times$} on \texttt{Qwen3-1.7B-Base} and \textbf{2.7$\times$} on \texttt{Qwen3-4B-Base}, outperforming all self-speculative decoding baselines. It also matches or surpasses the state-of-the-art EAGLE-3, despite EAGLE-3 relying on a separately trained drafter network. Beyond faster decoding, \methodname\ improves task performance over standard autoregressive (AR) fine-tuning, whereas training-based acceleration methods such as E2D2 and LayerSkip suffer severe quality degradation.

\subsection{Analysis}

Here we explain why \methodname\ outperforms early-exit and MTP-based self-speculative decoding methods as well as speculative decoding methods with a separate drafter like EAGLE-3. The advantages stem from both encoder and decoder: the encoder produces deeper, more future-predictive contextual representations, while the decoder, which shares parameters with the verifier, drafts in a way that remains closely aligned with the verifier's generation distribution. All experiments in this analysis, as well as the following ablation studies, are conducted by evaluating \texttt{Qwen3-1.7B-Base} models fine-tuned on GSM8K with the same setup as in our main experiments.

\textbf{Better Drafting from Deeper Representations and Parameter Sharing with the Verifier.\space}
We first show that conditioning on deep representations enables \methodname\ to outperform early-exit methods. As shown in Figure~\ref{fig:draft_tok_acc_rate_per_pos}, \methodname\ consistently achieves higher acceptance rates than LayerSkip despite using a much smaller $2$-layer drafter. Early-exit methods like LayerSkip terminate computation at intermediate layers and thus discard the deeper representations from later layers that are crucial for accurate token prediction. \methodname\ avoids this limitation by drafting from the very last layers of the model, giving the drafter access to the verifier's full-depth contextual features. Moreover, \methodname\ outperforms EAGLE-3 on draft acceptance even though EAGLE-3 also produces drafts conditioned on deep representations. We attribute this gain to parameter sharing: \methodname's drafter is the verifier's own last two layers, so each draft token is sampled from a distribution closely matching the verifier's. In contrast, EAGLE-3 relies on a separate draft model, whose output distribution can still deviate from the verifier's despite being trained to approximate it.

\textbf{Cheaper Drafting with A Lightweight Decoder.\space}
Figure~\ref{fig:draft_tok_acc_rate_per_pos} shows that \methodname\ and Apple MTP achieve comparable draft acceptance, since both condition drafting on deep contextual representations. The throughput gap therefore comes from drafting cost: MTP requires a full forward pass at every drafting step to produce the final hidden states, while \methodname\ reuses the deep contextual representations cached during verification, requiring only a 2-layer decoder pass per draft token. Appendix~\ref{sec:computational_cost} quantifies the dominant computation and gives an approximate latency model for fixed-length drafting.

\textbf{Improved Task Performance from Planning Ahead.\space}
We next explain the task performance gains of \methodname\ over standard AR fine-tuning. We hypothesize that our speculative drafting objective implicitly trains the encoder to produce more future-predictive contextual representations: because the decoder must accurately predict tokens many steps ahead from these representations, the encoder is pushed to encode information about future tokens. This is similar to the effect observed in MTP methods~\citep{gloeckle2024better, liu2024deepseek}. Our linear-probe experiment in Figure~\ref{fig:enc_hid_pred} directly tests this hypothesis: \methodname's encoder produces hidden representations that remain more predictive of distant future tokens than those of a standard AR model, consistent with a "planning-ahead" effect induced by the speculative drafting objective. Additional details on the experimental design are provided in Appendix~\ref{sec:enc_hid_pred}.

\begin{figure}[t]
\centering
\begin{minipage}{0.60\textwidth}
\centering
\begin{minipage}{0.49\linewidth}
    \centering
    \includegraphics[width=\linewidth]{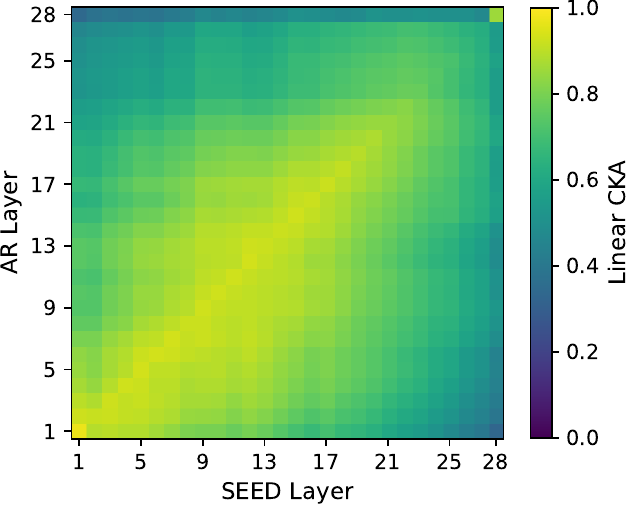}
\end{minipage}
\hfill
\begin{minipage}{0.49\linewidth}
    \centering
    \includegraphics[width=\linewidth]{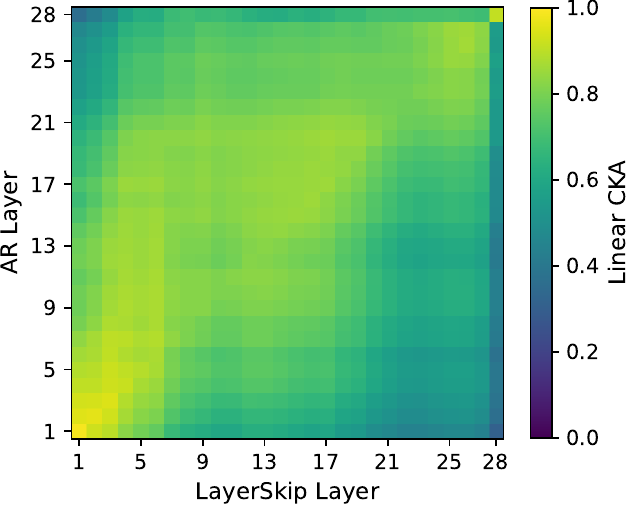}
\end{minipage}
\captionof{figure}{Cross-layer linear CKA matrices on GSM8K. \textbf{Left:} AR vs \methodname. \textbf{Right:} AR vs LayerSkip.}
\label{fig:CKA_heatmap}
\end{minipage}
\hfill
\begin{minipage}{0.38\textwidth}
\centering
\small
\captionof{table}{Ablation on the choice of $\lambda$. $l_\text{draft}$ and $r_\text{draft}$ denote the average draft token acceptance length and acceptance rate respectively.}
\label{tab:ablate_lambda}
\begin{tabular}{lcccc}
\toprule
$\lambda$ & \textbf{Acc.} & \textbf{Tput} & $l_\text{draft}$ ($r_\text{draft}$)  \\
\midrule
0.1 & 58.5 & 81.5 & 3.0 (87.4\%) \\
0.5 & 59.1 & 87.6 & 3.7 (88.3\%) \\
1.0 & 57.1 & 91.8 & 4.0 (88.6\%) \\
1.5 & 55.5 & 92.1 & 4.0 (88.0\%) \\
2.0 & 55.1 & 91.4 & 4.0 (88.4\%) \\
\bottomrule
\end{tabular}
\end{minipage}
\end{figure}

\textbf{What \methodname's Learned Representations Look Like.\space}
Finally, we examine the layer-wise structure of the representations learned by \methodname\ using Centered Kernel Alignment (CKA)~\citep{kornblith2019similarity}. With the hidden states collected from the linear-probe experiment, we compute cross-layer CKA matrices between the AR fine-tuned model and both the \methodname\ model and the LayerSkip model. Figure~\ref{fig:CKA_heatmap} shows that \methodname's representations remain very similar to the AR model's across the early and middle layers, indicating that the training dynamics of the encoder are largely unchanged relative to standard AR fine-tuning. The strongest divergence appears near the encoder-decoder interface (layer $l' = 26$), where \methodname\ reshapes representations most substantially, likely by encouraging the encoder to output more future-predictive hidden states and by adapting the decoder to operate on a novel mixture of raw token embeddings and cached deep representations. 

Together, these patterns suggest that \methodname's training acts as a \emph{localized} modification of the standard AR fine-tuning, targeting the layers responsible for drafting while preserving the deep contextual processing already performed by the bulk of the model. By contrast, LayerSkip alters the model's internal representations more substantially and shows broader representational drift across layers, explaining its degradation in downstream task performance.

\subsection{Ablation Study}
\label{sec:ablation}

\begin{figure}[h]
\centering
\begin{minipage}[t]{0.48\textwidth}
\centering
\vspace{0pt}
\includegraphics[width=\linewidth]{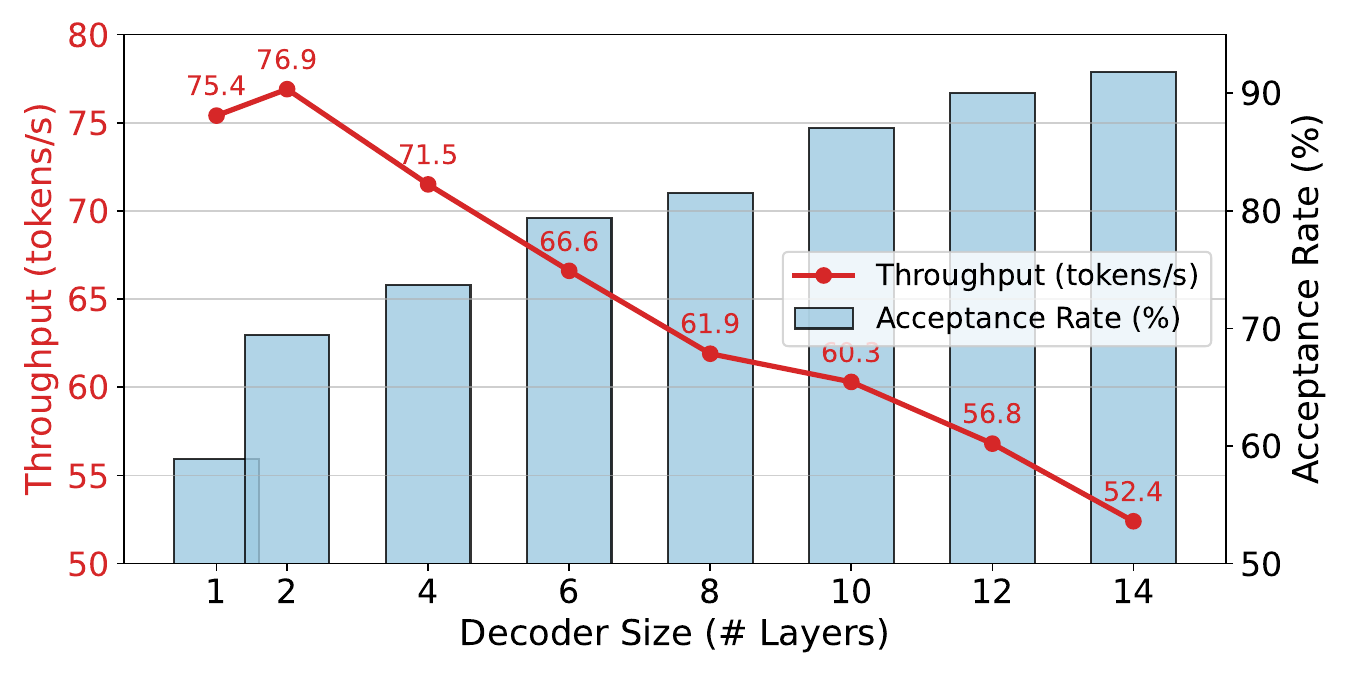}
\captionof{figure}{Ablation on decoder size. Models are evaluated with a simple fixed draft length of $4$, so that the draft token acceptance rate directly reflects draft quality and isolates how decoder size influences drafting accuracy. Larger decoders improve draft acceptance but reduce throughput due to higher drafting cost.}
\label{fig:ablate_decoder_size}
\end{minipage}
\hfill
\begin{minipage}[t]{0.48\textwidth}
\centering
\vspace{0pt}
\includegraphics[width=\linewidth]{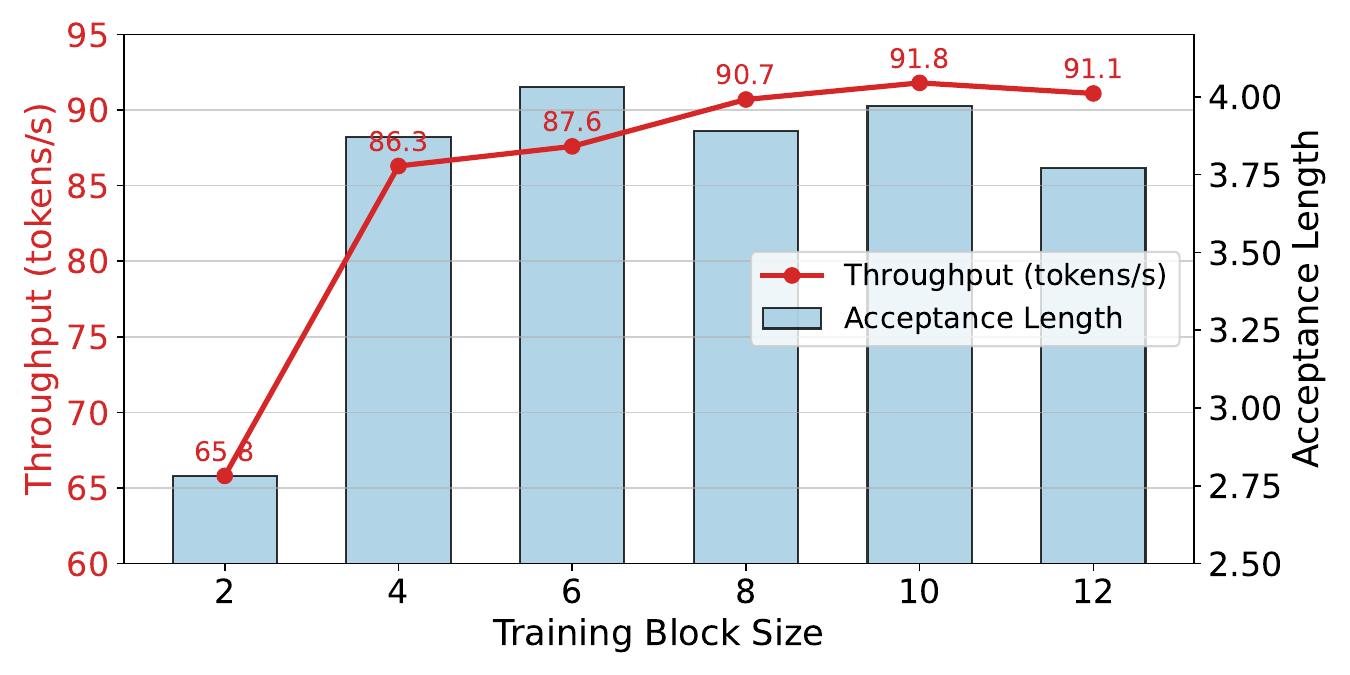}
\captionof{figure}{Ablation on training block size. Models are evaluated with dynamic drafting and tree-based verification. Larger block sizes improve draft acceptance and decoding throughput.}
\label{fig:ablate_training_block_size}
\end{minipage}

\end{figure}

Figure~\ref{fig:ablate_decoder_size} studies the effect of \textbf{decoder size}, where we keep the total number of transformer layers fixed and vary the encoder-decoder split (i.e., a larger decoder corresponds to a smaller encoder, and vice versa). Increasing the number of decoder layers improves draft quality, but it also increases the cost of each drafting step and reduces overall decoding throughput. For a better trade-off, we use only the last 2 layers as the decoder in our main experiments. Figure~\ref{fig:ablate_training_block_size} studies the effect of the \textbf{training block size} used in the speculative drafting objective $\mathcal{L}_{spec}$. Increasing the block size generally improves acceptance length and decoding throughput, suggesting that training the decoder to operate over longer draft spans better prepares it for speculative generation at inference time. We use a block size of 10 in our main experiments as it provides the best decoding speedup. Table~\ref{tab:ablate_lambda} studies the effect of the \textbf{loss weight $\lambda$} in Eq.~\eqref{eq:final_obj}. A larger $\lambda$ places more weight on $\mathcal{L}_{spec}$, generally improving draft quality, while a smaller $\lambda$ emphasizes the standard next-token prediction loss $\mathcal{L}_{CE}$, which favors task accuracy. We choose $\lambda = 1.0$ in our main experiments as it provides a favorable balance. Additional ablations are deferred to Appendix~\ref{sec:further_ablations}.

\section{Conclusion}
\methodname\ mitigates the central tradeoff in self-speculative decoding between draft quality and drafting cost by reframing decoder-only transformers as implicit encoder-decoder models. By reusing only the last layers as a lightweight drafter conditioned on cached deep representations of the verified prefix, \methodname\ achieves cheap drafting without discarding the rich contextual features produced by the preceding encoder. Across diverse benchmarks, \methodname\ consistently outperforms prior self-speculative baselines in decoding speed while preserving, and often improving, task performance relative to standard autoregressive fine-tuning. Our analysis further suggests that these gains arise because \methodname\ reshapes the training dynamics near the encoder-decoder interface, likely by encouraging the model to learn more future-predictive internal representations.

\section*{Acknowledgement}
This research is supported in part by NSF IIS-2508145, Amazon Research Award, and Lambda's Research Grant Program. We thank Ziteng Sun for his thoughtful comments on the manuscript. 

\newpage

\bibliographystyle{plain}
\bibliography{ref}

%%%%%%%%%%%%%%%%%%%%%%%%%%%%%%%%%%%%%%%%%%%%%%%%%%%%%%%%%%%%
\newpage
\appendix

\section{Limitations and Future Directions}
\label{sec:limitations_future_directions}
While \methodname\ achieves strong speedups and preserves task quality across the settings studied in this work, several limitations remain.

\paragraph{Scaling to Larger Models.}
Our experiments focus on \texttt{Qwen3-1.7B-Base} and \texttt{Qwen3-4B-Base} due to computational constraints. Although these models cover multiple tasks and already demonstrate consistent gains, we have not yet validated \methodname\ on larger models like 8B, 14B, or 32B-scale LLMs. The efficiency-quality tradeoff may change with model size: larger models have deeper stacks and potentially richer intermediate representations, which could make the encoder-decoder partition more effective. Evaluating \methodname\ at larger scales is therefore an important next step.

\paragraph{Beyond Task-Specific Fine-Tuning.}
In this paper, \methodname\ is mainly studied in task-specific supervised fine-tuning settings. An important direction for future work is extending \methodname\ beyond task-specific fine-tuning to larger-scale post-training or even pretraining settings. Applying the speculative drafting objective during general instruction tuning, preference optimization, continual post-training, or pretraining could produce models that are natively compatible with \methodname\ across a wider range of prompts and domains. Such a setting would also allow us to study whether the "planning-ahead" behavior induced by \methodname\ improves general model capabilities and robustness beyond the task-specific benchmarks considered here.

\paragraph{Adaptive Partitioning and Hardware-Aware Implementation.}
Our main experiments use a simple fixed partition, where the top two transformer layers serve as the decoder. This design is intentionally simple, but it is unlikely to be optimal for all model sizes, tasks, sequence lengths, or deployment environments. Future work could explore adaptive partitioning strategies that choose the encoder-decoder split based on model depth, layerwise representational quality, draft acceptance statistics, or runtime constraints. In addition, practical speedup depends heavily on implementation details such as KV-cache layout, tree-attention kernels, batching behavior, and GPU memory bandwidth. Hardware-aware implementations, including custom kernels for decoder-only drafting and tree-based verification, could further improve the realized speedup of \methodname\ in production serving systems.

\section{Additional Discussions on Related Work}
\label{sec:add_related_works}
A high-level comparison of representative LLM inference acceleration methods, including our approach, is summarized in Table~\ref{tab:baseline-comparison}. Unlike many prior methods, \methodname\ does not bring additional system complexity by introducing external modules, while achieving considerable inference speedups. More importantly, because \methodname\ preserves the standard next-token prediction objective and achieves generation quality on par with autoregression, it can be seamlessly integrated into existing supervised fine-tuning pipelines. This allows training data to simultaneously improve both inference efficiency and generation quality, an advantage not shared by most existing LLM acceleration methods, which typically leverage training data solely for inference speedup, often at the expense of generation quality. 

\begin{table}[h]
\centering
\caption{\textbf{Comparison of \methodname\ against existing LLM acceleration methods.} "Lossless" denotes no degradation in generation  quality compared to standard autoregressive models.}
\label{tab:baseline-comparison}
\vspace{0.5em}
\small
\renewcommand{\arraystretch}{1.2}
% \resizebox{\linewidth}{!}{
    \begin{tabular}{l c c c c}
    \toprule
    \textbf{Method} & \textbf{No Ext. Module} & \textbf{Lossless} & \textbf{Training Cost} & \textbf{Speedup} \\
    \midrule

    \textbf{General MTP} & & & & \\

    \makecell[l]{Meta MTP \citep{gloeckle2024better} \& \\ DeepSeek MTP \citep{liu2024deepseek}} & 
    \textbf{\xmark} & 
    \textbf{\cmark} & 
    High & 
    Mid \\

    FastMTP \citep{cai2025fastmtp} & 
    \textbf{\xmark} & 
    \textbf{\cmark} & 
    Low & 
    Mid \\

    Apple MTP \citep{samragh2025your} & 
    \textbf{\xmark} & 
    \textbf{\cmark} & 
    Low & 
    High \\

    Jacobi Forcing \citep{hu2025fast} & 
    \textbf{\cmark} & 
    \textbf{\xmark} & 
    Low & 
    High \\

    \midrule

    \textbf{Diffusion Language Models} & & & & \\

    E2D2 \citep{arriola2025encoder} & 
    \textbf{\cmark} & 
    \textbf{\xmark} &
    Low & 
    Mid \\

    TiDAR \citep{liu2025tidar} & 
    \textbf{\cmark} & 
    \textbf{\xmark} & 
    High & 
    High \\

    \midrule

    \textbf{General Speculative Decoding} & & & & \\

    MEDUSA \citep{cai2024medusa} & 
    \textbf{\xmark} & 
    \textbf{\cmark} & 
    Low & 
    High \\
    
    EAGLE-3 \citep{li2025eagle3} & 
    \textbf{\xmark} & 
    \textbf{\cmark} & 
    High & 
    High \\

    \midrule

    \textbf{Self-Speculative Decoding} & & & & \\

    LayerSkip \citep{elhoushi2024layerskip}& 
    \textbf{\cmark} & 
    \textbf{\xmark} & 
    Low & 
    Low \\
    
    \makecell[l]{SWIFT \citep{xia2025swift} \& \\ CLaSp \citep{chen2025clasp}} & 
    \textbf{\cmark} & 
    \textbf{\cmark} & 
    None & 
    Low \\

    CAS-Spec \citep{ning2025cas} & 
    \textbf{\cmark} & 
    \textbf{\cmark} & 
    None & 
    Mid \\

    \methodname~(Ours) & 
    \textbf{\cmark} & 
    \textbf{\cmark} & 
    Low &
    High \\

    \bottomrule
    \end{tabular}
% }
\end{table}

\subsection{Connection to Self-Speculative Decoding with Layer Skipping}
\label{para:comparison_with_layerskip}

A closely related approach to \methodname\ is self-speculative decoding with \emph{layer skipping}, which accelerates drafting by omitting a subset of transformer layers during the draft forward pass \citep{liu2024kangaroo, xia2025swift, zhang2024draft}. 
Using the notation from Section~\ref{sec:methodology}, this strategy can also be interpreted within our encoder–decoder framework. Specifically, the skipped layers $\mathcal{E} = \{e_1, e_2, \dots, e_s\}$ can be viewed as the encoder, while the remaining layers $\mathcal{D} = \{d_1, \dots, d_{l-s}\}$ act as the decoder that generates draft tokens.

However, this formulation reveals an inherent limitation of layer-skipping approaches. Suppose we have some verified context tokens $X_1, X_2, \dots, X_n$ and are about to propose a new draft token. At any decoder layer $\ell \in \mathcal{D}$, the attention mechanism
\begin{align*}
\text{Attn}(\hiddenreps{n}{\ell}, \kvcache{< n}{\ell})
\end{align*}
operates on representations at the same layer, where verified tokens ($i < n$) are represented by their cached keys and values $\kvcache{< n}{\ell}$ obtained during the previous verification step. In typical layer-skipping schemes, a large portion of the retained layers $\mathcal{D}$ include early layers of the transformer. Consequently, the KV caches available to these layers correspond to relatively shallow representations of the verified prefix. The draft model therefore attends only to low-level contextual features when generating candidate tokens, which can significantly degrade draft quality.

In contrast, \methodname\ places the drafting component strictly in the \emph{last few} layers of the transformer ($\ell \in [l'+1, l]$). 
Rather than conditioning on a single encoder-produced hidden state, the decoder attends directly to the verified KV caches $\kvcache{\le n}{dec}$, which were computed after the verified tokens passed through all $l'$ encoder layers. These cached keys and values therefore encode the deepest contextual representations available in the model and effectively serve as the shared encoder state.

As a result, the draft model in \methodname\ always conditions on deep, highly contextualized representations of the verified prefix. 
This asymmetric design preserves the representational strength of the full model while keeping the draft computation inexpensive, enabling higher draft acceptance rates and improved overall decoding throughput.

\subsection{Comparison to the EAGLE Series}
\label{sec:eagle1_comp}

Here we provide a deeper theoretical and architectural comparison between \methodname\ and the EAGLE series of speculative decoding algorithms \citep{li2024eagle1, li2024eagle2, li2025eagle3}. Both approaches share a fundamental motivation: draft models can achieve significantly higher acceptance rates if they are conditioned on deep, highly structured contextual representations of the verified prefix, rather than shallow representations or raw tokens alone. However, the methods diverge in how they formulate the drafting objective and system architecture.

The original EAGLE \citep{li2024eagle1} formulates the draft model as performing feature-level autoregression. Given a sequence of hidden representations $H_0, H_1, \dots, H_{n-1}$, the draft model learns to predict the next hidden representation $H_n$. Because continuous hidden representations are highly structured, this regression task becomes easier to learn than directly predicting raw discrete tokens. However, conditional dependencies in language are based on \emph{discrete identity}. To resolve this, EAGLE samples the discrete token $X_{n+1}$ from $H_n$ and then fuses the embedding of $X_{n+1}$ with $H_n$ to predict the next step. Yet one problem remains: during training, EAGLE's draft model receives ground-truth features from the target model as input at every position, but at inference time, the draft model must condition on its own predicted features for speculative positions. This creates an inherent exposure bias and distribution shift, because if the draft model produces poor feature predictions early on, subsequent inputs will be severely out-of-distribution. This motivates the authors in EAGLE-3 \citep{li2025eagle3} to drop feature regression entirely and instead focus purely on token prediction, handling the exposure bias through a training-time drafting mechanism.

Meanwhile, \methodname\ operates under the simplified hypothesis that hidden representations are valuable not because they are easier to predict, but because they make predicting the next token easier. By formulating drafting as an implicit encoder-decoder process, \methodname\ avoids the feature-level distribution shift entirely. During both training and inference, the decoder processes draft tokens using only their raw token embeddings ($H^0$), while cross-attending to the deep contextual representations ($KV^{dec}$) of the verified prefix. Because the decoder is explicitly trained to handle this specific mixture of representation types, it never has to condition on its own continuous feature predictions. Consequently, \methodname\ achieves high draft quality through direct token prediction, without the need for auxiliary feature regression objectives or feature fusion mechanisms. Moreover, because the drafter shares parameters with the verifier, it remains better aligned with the target model's distribution. In contrast to EAGLE-3, which conditions drafting on multiple levels of intermediate features and requires a complex training-time test simulation to mitigate exposure bias, \methodname\ proves that it is possible to achieve higher draft quality by simply reusing the verifier's high-level contextual representations through the KV cache.

\subsection{Comparison to E2D2}
\label{sec:e2d2_comp}

E2D2~\citep{arriola2025encoder} pioneered the amortization principle of separating expensive clean-context encoding from repeated lightweight generation. \methodname\ can be viewed as bringing this principle into autoregressive decoding with exact verifier-based acceptance. Concretely, the new components are: (i) the speculative drafting objective that trains the last layers to draft from raw token embeddings while conditioning on verifier-cached deep KV representations, (ii) the fully block-causal training masks that preserve the verifier as a valid autoregressive model, and (iii) the resulting draft-then-verify inference loop in which verification and encoding merge into a single forward pass. Crucially, this AR speculative formulation changes the nature of the result: instead of merely extending the quality–efficiency frontier, exact verification enables acceleration without sacrificing model performance.

To disentangle whether this advantage comes from the formulation or from implementation / training choices, we run a controlled comparison against E2D2 on GSM8K with \texttt{Qwen3-1.7B-Base}, matching the training budget, training block size $b=4$, and encoder-decoder split, and disabling \methodname's adaptive drafting and tree verification with a fixed draft length of 4. This matches the split and generation span, while the inference procedures retain their respective diffusion and speculative computations.

\begin{table}[htbp]
\centering
\caption{Controlled comparison with E2D2. We compare E2D2 and \methodname\ across different decoder sizes by training and evaluating corresponding \texttt{Qwen3-1.7B-Base} models on GSM8K. \methodname\ achieves much better task performance while using a smaller decoder, proving that exact verification enables acceleration without sacrificing model performance.}
\vspace{0.5em}
\label{tab:e2d2_controlled}
\small
\begin{tabular}{ccccc}
\toprule
\makecell{Decoder Size\\(\# layers)} & \makecell{E2D2\\Acc. (\%)} & \makecell{E2D2\\Tput (tokens/s)} & \makecell{\methodname\\Acc. (\%)} & \makecell{\methodname\\Tput (tokens/s)} \\
\midrule
2 & 23.7 & 75.9 & 57.5 & 74.3 \\
4 & 26.6 & 67.6 & 57.9 & 66.1 \\
6 & 32.3 & 63.2 & 59.4 & 60.8 \\
8 & 32.4 & 57.3 & 58.8 & 56.3 \\
10 & 36.3 & 51.2 & 57.4 & 53.3 \\
12 & 41.3 & 46.4 & 57.0 & 48.8 \\
14 & 43.4 & 42.8 & 57.6 & 46.4 \\
\bottomrule
\end{tabular}
\end{table}

The two methods achieve nearly identical throughput at every split, confirming that they share the same amortization mechanism. The difference is in task quality: E2D2 must trade accuracy for speed, whereas \methodname's accuracy is essentially flat across all decoder sizes because the exact verifier guarantees that the output distribution matches autoregressive decoding regardless of drafter capacity. This lets us use the smallest, fastest decoder while securing both speed and accuracy.

\section{Further Details on Experimental Setup}
\label{sec:further_exp_setup}
\subsection{Hardware}
\label{sec:hardware}
All 1.7B models are fine-tuned on a single 48 GB NVIDIA A6000 GPU, and all 4B models are fine-tuned on either NVIDIA 80 GB H100 GPUs or 180 GB B200 GPUs. Some 4B models are trained on two H100 GPUs using PyTorch Fully Sharded Data Parallel (FSDP)~\citep{zhao2023pytorch}, while others are trained on a single B200 GPU. All trained models are evaluated on a single NVIDIA A6000 GPU to measure task performance and throughput.

\subsection{Datasets}
\label{sec:datasets_details}
Following E2D2~\citep{arriola2025encoder}, we train and evaluate task-specific models on four datasets covering math reasoning, code generation, general knowledge, and summarization: GSM8K~\citep{cobbe2021training}, KodCode~\citep{xu2025kodcode}, ScienceQA~\citep{lu2022learn}, and CNN/Daily Mail~\citep{nallapati2016abstractive}.

For GSM8K, we use the Hugging Face dataset \texttt{openai/gsm8k}, with the official train split for training and test split for evaluation. We pre-process the data by adding a prefix \texttt{"Please reason step by step, and put your final answer within \$\textbackslash boxed\{\}\$."} to the inputs, prefacing the answers with \texttt{"Answer: "}, and wrapping the solutions in \texttt{"\$\textbackslash boxed\{\}\$"}. Inputs and targets are truncated to a maximum length of $384$ tokens each, ensuring a maximum sequence length of $768$ during training. Final evaluation is performed in $0$-shot mode using the \texttt{lm-eval harness} library with the "flexible match" criteria, and similar pre-processing is applied to question texts. Post-processing is done to truncate text at \texttt{<|endoftext|>} tokens, and solutions are reverted to their original form of \texttt{"\#\#\# <Answer>"}. Inputs are pre-processed as above.

For KodCode, we use the Hugging Face dataset \texttt{KodCode/KodCode-V1-SFT-R1} with a deterministic reconstruction of train/test splits. We first keep only samples with online judge style questions, then randomly subsample $50\%$ of these examples, and finally define the test set as the last $1000$ samples of this subsample while using the remainder as the training set. Each problem is prompted as a Python programming task ending with a \texttt{[BEGIN]} marker. Inputs and targets are truncated to a maximum length of $512$ tokens each, ensuring a maximum sequence length of $1024$ during training. During evaluation, we restrict the test set to problems marked with difficulty level "easy". For each problem, the model generates a program which is then executed against the provided test cases, and we report exact problem-solving accuracy based on whether all tests pass.

For ScienceQA, we use the Hugging Face dataset \texttt{derek-thomas/ScienceQA} and restrict it to text-only examples. The training set is constructed by merging the original train and validation splits, while the evaluation set is created by sampling $1000$ examples from the original test split. Inputs are formatted as multiple-choice questions with answer options labeled (A) to (H), and targets consist of the reasoning followed by a canonical final statement of the form \texttt{"The answer is (X)"}. Inputs and targets are truncated to a maximum length of $384$ tokens each, ensuring a maximum sequence length of $768$ during training. For evaluation, the predicted option letter is extracted from the generated text, and exact-answer accuracy is reported.

For CNN/Daily Mail, we use the Hugging Face dataset \texttt{abisee/cnn\_dailymail} (version 3.0.0), with the official train split for training and the test split for evaluation. Inputs consist of the news article prefixed with a summarization instruction, while targets are the corresponding summaries prefixed with \texttt{"Summary: "}. Unlike the previous datasets, we allocate the token budget asymmetrically, assigning $90\%$ of the sequence length to the article and $10\%$ to the summary, for a total sequence length of $1024$ tokens. During evaluation, we apply the same length constraints and evaluate on up to $1000$ samples from the test split. Summarization quality is measured using ROUGE-1, ROUGE-2, and ROUGE-L.

\subsection{Baselines}
\label{sec:baseline_details}
Our overall implementation builds on the official E2D2 codebase\footnote{\url{https://github.com/kuleshov-group/e2d2}}~\citep{arriola2025encoder}. Accordingly, our AR and E2D2 baselines follow their implementation. For E2D2, we use the last $4$ layers as the decoder for a better trade-off between performance and speedup. Consistent with the original setup, E2D2 is trained and evaluated using a block size of $4$.

For Apple MTP~\citep{samragh2025your}, the original paper does not provide an official code release. We therefore adopt the open-source unofficial implementation from \url{https://github.com/siihwanpark/MTP-GLoRA} and make the necessary modifications to better match the algorithmic details described in the paper. We fine-tune the model to handle $8$ mask tokens as input by training gated-LoRA weights and a $2$-layer MLP sampler module, using a LoRA rank of $16$ and a LoRA alpha of $32$. For evaluation, we set the draft length to $8$ across all datasets, with the exception of CNN/Daily Mail, where a draft length of $4$ is used.

For LayerSkip~\citep{elhoushi2024layerskip}, the released repository provides inference code but not training code. We therefore use the implementation in Hugging Face TRL\footnote{\url{https://github.com/huggingface/trl/pull/3111}} and make the necessary adjustments to align it with the training procedure described in the original paper. The early-exit layer is set to layer $8$ for \texttt{Qwen3-1.7B-Base} and layer $12$ for \texttt{Qwen3-4B-Base}. We fine-tune the base models using a composite objective $L = L_\text{final} + \lambda \cdot L_\text{early-exit}$ with $\lambda=1.0$. Training follows a rotational early-exit curriculum with stride $8$, and layer dropout is applied with a constant schedule and a maximum dropout rate of $0.1$.

For EAGLE-3~\citep{li2025eagle3}, we use the publicly available AngelSlim checkpoints \texttt{AngelSlim/Qwen3-} \texttt{1.7B\_eagle3}\footnote{\url{https://huggingface.co/AngelSlim/Qwen3-1.7B_eagle3}} and \texttt{AngelSlim/Qwen3-4B\_eagle3}\footnote{\url{https://huggingface.co/AngelSlim/Qwen3-4B_eagle3}} paired with \texttt{Qwen3-1.7B} and \texttt{Qwen3-4B} instruct models due to the lack of corresponding EAGLE-3 weights for the base models. Evaluation is conducted using the official EAGLE GitHub repository\footnote{\url{https://github.com/SafeAILab/EAGLE}}. Note that although EAGLE-3 reports more than $5\times$ average speedup on Vicuna-, LLaMA-, and DeepSeek-based models, our experiments on \texttt{Qwen3-1.7B} and \texttt{Qwen3-4B} show only around $2\times$ speedup on a single NVIDIA A6000. This observation is consistent with the results reported by AngelSlim in \url{https://huggingface.co/AngelSlim/Qwen3-1.7B_eagle3}.

For PARD~\citep{an2026pard}, we evaluate the released target-independent Qwen3 drafter\footnote{\url{https://huggingface.co/amd/PARD-Qwen3-0.6B}}, paired with the corresponding task-specific, fine-tuned \texttt{Qwen3-4B-Base} AR checkpoint under the same evaluation setting as \methodname. We include PARD only in the 4B portion of Table~\ref{tab:main-results}, as its 0.6B-parameter drafter is relatively large for a 1.7B target model.

For SWIFT~\citep{xia2025swift} and DEL~\citep{zarch2025context}, we use the authors' released implementations directly. SWIFT runs on top of our fine-tuned AR checkpoints, with Bayesian optimization interval $1$, update interval $25$, layer skip ratio $0.4$, context window $50$, maximum optimization iterations $1000$, tolerance limit $300$, and early-stop score threshold $0.95$. DEL runs on top of our fine-tuned LayerSkip checkpoints by maintaining an exponential moving average on acceptance/confidence statistics with retention factor $\omega=0.95$.

\subsection{Training}
\label{sec:training_details}

We train all models using the MosaicML Composer framework\footnote{\url{https://docs.mosaicml.com/projects/composer}}. Unless otherwise noted, we use the same optimization setup across methods to ensure a fair comparison. Specifically, we apply a linear learning-rate warm-up for the first $100$ steps, followed by cosine decay. Model checkpoints are selected based on validation loss with early stopping. We set the random seed to $1$ during training.

Training hyperparameters vary by model scale and dataset, primarily in learning rate, batch size, and maximum sequence length. Table~\ref{tab:training_hparams} summarizes the final training setup.

\begin{table}[h]
\centering
\caption{\textbf{Training configurations used for task-specific fine-tuning.} For each dataset and model scale, all methods share the same optimizer schedule and training budget to maintain comparability.}
\label{tab:training_hparams}
\vspace{0.5em}
\begin{tabular}{lccccc}
\toprule
\textbf{Dataset} & \textbf{Base Model} & \textbf{Batch Size} & \textbf{LR} & \textbf{Steps} & \textbf{Max Length} \\
\midrule
GSM8K & \texttt{Qwen3-1.7B-Base} & $1$ & $1e^{-5}$ & $30{,}000$ & $768$ \\
KodCode & \texttt{Qwen3-1.7B-Base} & $1$ & $1e^{-5}$ & $30{,}000$ & $1024$ \\
ScienceQA & \texttt{Qwen3-1.7B-Base} & $1$ & $1e^{-5}$ & $30{,}000$ & $768$ \\
CNN/Daily Mail & \texttt{Qwen3-1.7B-Base} & $32$ & $1e^{-5}$ & $30{,}000$ & $1024$ \\
GSM8K & \texttt{Qwen3-4B-Base} & $1$ & $5e^{-6}$ & $30{,}000$ & $768$ \\
KodCode & \texttt{Qwen3-4B-Base} & $1$ & $5e^{-6}$ & $30{,}000$ & $1024$ \\
ScienceQA & \texttt{Qwen3-4B-Base} & $1$ & $5e^{-6}$ & $30{,}000$ & $768$ \\
CNN/Daily Mail & \texttt{Qwen3-4B-Base} & $32$ & $5e^{-6}$ & $30{,}000$ & $1024$ \\
\bottomrule
\end{tabular}
\end{table}

\subsection{Task Prompts}

We now describe the task prompts used to train and evaluate the baseline methods in our experiments. The prompts are designed to be simple and straightforward, as detailed below:

\begin{promptbox}{Task Prompts}

\textbf{GSM8K}

\vspace{0.5em}

Please reason step by step, and put your final answer within \$\textbackslash boxed\{\}\$. \{question\}

\vspace{0.5em}

\textbf{KodCode}

\vspace{0.5em}

You are an expert Python programmer. Solve the following problem.

\vspace{0.5em}

\{question\}

\vspace{0.5em}

\textbf{ScienceQA}

\vspace{0.5em}

The following is a multiple choice question. Think step by step and then give your final answer.

\vspace{0.5em}

\{question\}

\vspace{0.5em}

(A) \{choice A\}  

(B) \{choice B\}

...

\vspace{0.5em}

\textbf{CNN/Daily Mail}

\vspace{0.5em}

Summarize the following article: \{article\}

\end{promptbox}

\subsection{Evaluation}

We evaluate all trained models using greedy decoding. To measure decoding throughput, we adopt a different protocol from prior work, which typically continues generation until the maximum sequence length is reached. We find that if we force models to continue generating after the \texttt{<|endoftext|>} token, draft models tend to adapt to repetitive output patterns, which can artificially increase drafting accuracy and decoding throughput. However, in real-world applications, we only care about the tokens generated before the \texttt{<|endoftext|>} token. Therefore, in all experiments, we measure decoding throughput by terminating generation as soon as the \texttt{<|endoftext|>} token is generated.

\section{\methodname\ Implementation Details}
\label{sec:impl_details}
\subsection{Algorithms}
\label{sec:algorithms}

\begin{figure*}[t]
    \centering
    \includegraphics[width=\textwidth]{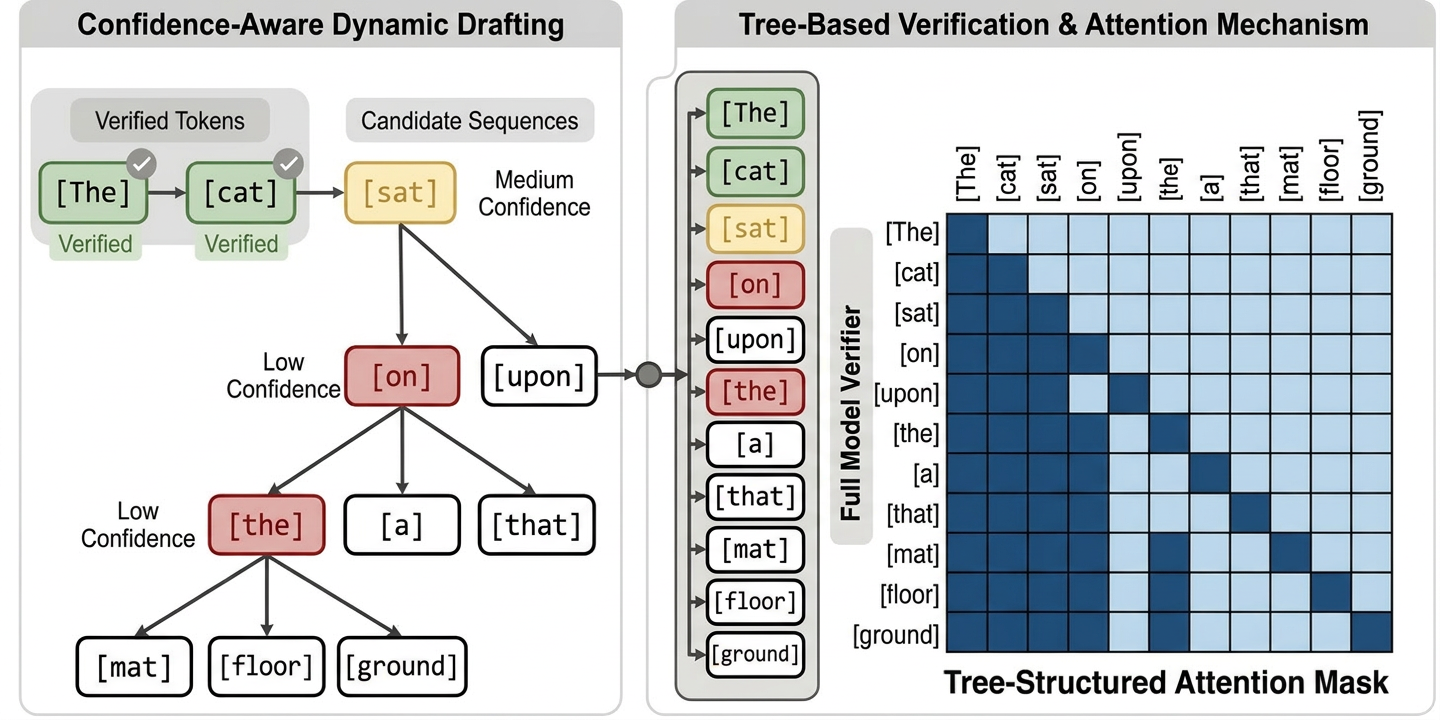}
    \caption{Illustration of confidence-aware dynamic drafting and tree-based verification in \methodname. \textbf{Left:} starting from a verified prefix, the lightweight decoder drafts autoregressively, with the current node expanded into multiple candidates using a top-$p$ rule, thus producing a small tree of candidate continuations. \textbf{Right:} all candidate tokens are packed and verified in one full-model forward pass using a tree-structured attention mask. Each candidate token attends to the entire verified prefix and to the candidate tokens on its own ancestral path, but not to tokens from unrelated branches. This allows parallel verification of multiple draft branches while preserving the causal dependencies of autoregressive decoding.}
    \label{fig:dynamic-tree-verification}
\end{figure*}

We include \methodname\ training and inference with fixed draft length in Algorithm~\ref{alg:dual-decode-train-alg} and Algorithm~\ref{alg:inference-alg}. Since our main experiments use adaptive drafting with tree-based verification, we further describe that procedure here and illustrate it in Figure~\ref{fig:dynamic-tree-verification}.

\begin{algorithm}
\caption{\methodname\ Training}
\begin{algorithmic}[1]
\REQUIRE Transformer $f = \decoder \circ \encoder$, encoder-decoder divider $l'$, training data $D$, block-size $b$, speculative drafting objective weight $\lambda$
\FOR{$X \in D$}
\STATE $H^0 \gets W_{in} X$ \tikzmark{top_brace}
\STATE $H^{l'} \gets \encoder(X)$
\STATE \LineComment{Standard next-token prediction loss}
\STATE $\mathcal{L}_{CE} \gets \sum_{j=2}^{|X|} - \log \decoder(X_j \mid H_{j-1}^{l'}, \kvcache{<(j-1)}{dec})$ \tikzmark{bot_brace}
% \STATE $n' = b$, $n = |X|$
\STATE\LineComment{In practice, this for loop can be computed in parallel for all blocks using the block causal attention mask illustrated in Figure~\ref{fig:train-attn-mask}}
\STATE $L_{spec} \gets 0$
% \WHILE{$n' \leq n$}
\FOR{$j=2$ to $|X|$}
\STATE \LineComment{Compute the start index of the block to which the last input token $X_{j-1}$ belongs}
\STATE $n' \gets \lfloor \frac{j-2}{b} \rfloor \times b + 1$
\STATE \LineComment{Speculative drafting objective}
\STATE $\mathcal{L}_{spec} \gets \mathcal{L}_{spec} - \log \decoder \left(X_j \mid \hiddenreps{n' \le \cdot \le (j-1)}{0}, \kvcache{<n'}{dec} \right)$
\ENDFOR
\STATE $L = \frac{\mathcal{L}_{CE} + \lambda \mathcal{L}_{spec}}{1 + \lambda}$
\STATE $\nabla \theta \gets \text{Backprop}(\mathcal{L})$
\STATE $f \gets \text{OptimizerStep}(f, \nabla \theta)$
\ENDFOR 
\STATE return $f$
\end{algorithmic}
\label{alg:dual-decode-train-alg}
\begin{tikzpicture}[remember picture, overlay]
    % Define the coordinates from the markers safely
    \coordinate (top) at (pic cs:top_brace);
    \coordinate (bot) at (pic cs:bot_brace);
    % Draw the brace
    \draw[decorate, decoration={brace, amplitude=4pt}, thick, algcomment]
        ([yshift=2ex, xshift=1.5em]top -| bot) 
        -- 
        ([yshift=-1ex, xshift=1.5em]bot)
        node[midway, right=0.5em, align=left, font=\footnotesize] 
        {Computed in a single forward pass of $f$};
\end{tikzpicture}
\end{algorithm}
\begin{algorithm}
\caption{\methodname\ Inference with Fixed Draft Length}
\begin{algorithmic}[1]
\REQUIRE Transformer with encoder $\encoder = f_{1:l'}$, decoder $\decoder = f_{l'+1:l}$, prompt $X_{1:t}$, draft length $d$, target length $T$
\STATE Initialize $KV_{1:t}$ by calling $(\decoder \circ \encoder)(X_{1:t})$
\WHILE{$t < T$}
    \STATE \LineComment{Drafting Phase: Autoregressively draft $d$ tokens using only the decoder}
    \STATE $X_{draft}[0] \gets X_t$
    \FOR{$i = 1$ to $d$}
        \STATE $q_i(x) \gets \decoder(W_{in}X_{draft}[i-1], KV_{1:(t+i-2)}^{dec})$
        \STATE $X_{draft}[i] \gets \arg \max q_i(x)$
    \ENDFOR

    \STATE $KV_{1:(t-1)} \gets \text{Truncate}(KV_{1:(t+d-1)}, t-1)$
    
    \STATE \LineComment{Verification Phase: Forward the full model in parallel}
    \STATE $p_1(x), \ldots, p_{d+1}(x) \gets (\decoder \circ \encoder)(X_{draft}[0:d], KV_{1:(t-1)})$

    \STATE $n_{accept} \gets 0$
    \FOR{$i = 1$ to $d$}
        \IF{$\arg\max_x p_i(x) = X_{draft}[i]$}
            \STATE $n_{accept} \gets n_{accept} + 1$
        \ELSE
            \STATE \textbf{break}
        \ENDIF
    \ENDFOR
    
    \STATE \LineComment{Append verified tokens and a bonus correction token}
    \STATE $x_{new} \gets \arg\max_x p_{n_{accept}+1}(x)$
    \STATE $X_{(t+1):(t+n_{accept}+1)} \gets [X_{draft}[1:n_{accept}], x_{new}]$
    \STATE $KV_{1:(t+n_{accept})} \gets \text{Truncate}(KV_{1:(t+d)}, t+n_{accept})$
    \STATE $t \gets t + n_{accept} + 1$
\ENDWHILE
\STATE \textbf{return} $X_{1:T}$
\label{alg:inference-alg}
\end{algorithmic}
\end{algorithm}

\paragraph{Confidence-Aware Dynamic Drafting.}
A fixed draft length is often suboptimal because the decoder's confidence varies across generation steps. Instead of always drafting exactly $d$ tokens, we let the decoder continue drafting as long as its confidence remains above a predefined threshold. Starting from the current verified prefix, the decoder autoregressively proposes the next token while conditioning on the verified decoder KV cache and the raw embeddings of previously drafted tokens, exactly as in the fixed-length setting. Also, at each drafting step, we expand the current node into multiple sibling candidates using a top-$p$ rule, following prior adaptive self-speculative decoding methods. Repeating this procedure over several steps produces a small tree of candidate continuations rather than a single draft sequence. This confidence-aware strategy allocates computation more effectively. When the decoder is confident, generation proceeds with a narrow branch and low drafting overhead. When uncertainty increases, the procedure broadens the candidate set so that the verifier can evaluate several plausible continuations in parallel. In practice, we set $\tau=0.7$ for GSM8K, KodCode, and ScienceQA, and $\tau=0.5$ for CNN/Daily Mail. For candidate expansion under the top-$p$ rule, let $p_0$ denote the probability of the current top-1 draft token. We then choose the number of candidate draft tokens $k$ at each step according to

\begin{align*}
    k =
    \begin{cases}
    1 & \text{if } p_0 > 0.95, \\
    3 & \text{if } 0.8 < p_0 \leq 0.95, \\
    5 & \text{if } 0.5 <p_0 \leq 0.8, \\
    10 & \text{otherwise.}
    \end{cases}
\end{align*}

\paragraph{Tree-Based Verification.}
Once the candidate tree is constructed, we verify all drafted tokens with a single full-model forward pass. To do so, we linearize the tree nodes into a packed sequence and apply a tree-structured attention mask that preserves the correct autoregressive dependencies. Concretely, every candidate token is allowed to attend to: (i) all verified prefix tokens, and (ii) the candidate tokens on its own ancestral path in the draft tree. Tokens from different branches are not allowed to attend to one another unless they share the same ancestors. This reproduces exactly the causal structure each branch would have under independent autoregressive evaluation, while enabling all candidates to be processed in parallel in one verifier call. After the verifier produces logits for all candidate nodes, we traverse the tree from the root and greedily accept the longest prefix whose verified predictions match the drafted tokens. As in standard speculative decoding, after the accepted draft prefix is determined, we additionally append one bonus token predicted by the verifier at the first non-accepted position. The KV cache is then truncated to the accepted path, the verified prefix is updated, and the next round of drafting begins.

\subsection{Further Ablations}
\label{sec:further_ablations}

\begin{figure*}[h]
    \centering
    \includegraphics[width=\textwidth]{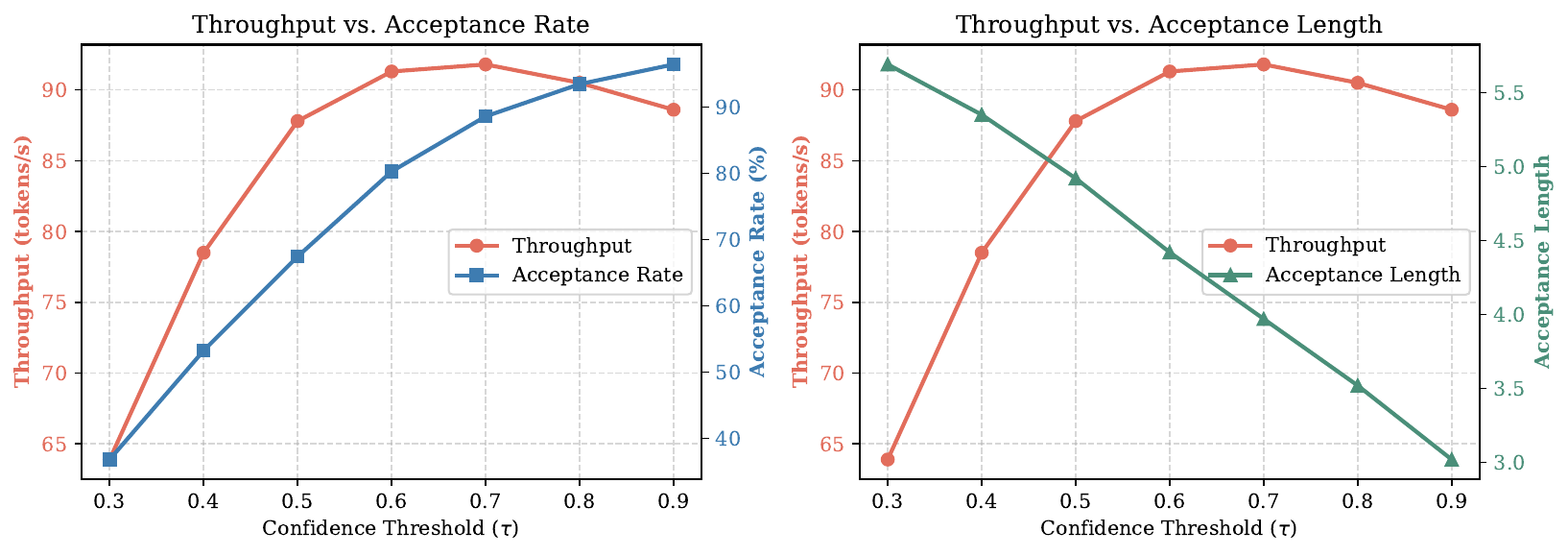}
    \caption{Effect of the confidence threshold $\tau$ used in dynamic drafting. We evaluate a GSM8K-finetuned \texttt{Qwen3-1.7B-Base} \methodname\ model on GSM8K and report average decoding throughput, draft token acceptance rate, and draft token acceptance length. Increasing $\tau$ improves acceptance rate but shortens acceptance length due to earlier termination of drafting. Overall decoding throughput remains stable for $\tau$ between $0.5$ and $0.9$, indicating that \methodname\ is relatively insensitive to the precise choice of $\tau$.}
    \label{fig:ablate_conf_threshold}
\end{figure*}

\paragraph{Effect of the Confidence Threshold $\tau$.}
In our main experiments, \methodname\ employs a dynamic drafting strategy that terminates drafting when the top-$1$ probability of the draft token falls below a confidence threshold $\tau$. To study the impact of $\tau$ on decoding efficiency, we vary the threshold and measure the resulting decoding throughput, draft token acceptance rate, and average acceptance length. Specifically, we evaluate a GSM8K-finetuned \texttt{Qwen3-1.7B-Base} \methodname\ model on GSM8K, and the results are shown in Figure~\ref{fig:ablate_conf_threshold}. As $\tau$ increases, the acceptance rate of draft tokens consistently improves, indicating that the internal confidence of the draft model correlates positively with draft quality. However, a higher $\tau$ also shortens the average acceptance length because drafting terminates earlier. Despite this trade-off, decoding throughput remains relatively stable within a wide range of thresholds ($\tau \in [0.5, 0.9]$), demonstrating that \methodname\ is robust to the choice of $\tau$. In our experiments, we select $\tau = 0.7$ as it provides a favorable balance between acceptance rate and drafting length. For more challenging generation tasks such as CNN/Daily Mail summarization, where draft token acceptance lengths are typically shorter, we adopt a slightly lower threshold ($\tau = 0.5$) to maintain longer drafting segments and better efficiency.

\begin{table}[h]
\centering
\caption{Ablation study on tree-based verification. We evaluate a GSM8K-finetuned \texttt{Qwen3-1.7B-} \texttt{Base} \methodname\ model on GSM8K. We report decoding throughput (Tput) in tokens/s, average draft token acceptance length (Acc. Len), and draft token acceptance rate (Acc. Rate) under fixed-length and dynamic drafting strategies. Tree-based verification consistently improves acceptance statistics and overall throughput.}
\label{tab:tree_attn_ablation}
\vspace{0.5em}
\resizebox{\textwidth}{!}{
\begin{tabular}{l ccc ccc ccc}
\toprule
\multirow{2}{*}{\textbf{Method}} & \multicolumn{3}{c}{\textbf{Fixed-Length Drafting ($d=4$)}} & \multicolumn{3}{c}{\textbf{Fixed-Length Drafting ($d=8$)}} & \multicolumn{3}{c}{\textbf{Dynamic Drafting ($\tau=0.7$)}} \\
\cmidrule(lr){2-4} \cmidrule(lr){5-7} \cmidrule(lr){8-10}
 & Tput & Acc. Len & Acc. Rate & Tput & Acc. Len & Acc. Rate & Tput & Acc. Len & Acc. Rate \\
\midrule
w/ Tree Attn  & 82.3 & 3.16 & 79.80\% & 82.5 & 4.65 & 59.10\% & 91.8 & 3.97 & 88.57\% \\
w/o Tree Attn & 76.9\drop{5.4} & 2.67 & 69.45\% & 75.2\drop{7.3} & 3.67 & 46.64\% & 85.1\drop{6.7} & 3.28 & 77.59\% \\
\bottomrule
\end{tabular}
}
\end{table}

\paragraph{Effect of Tree-Based Verification.} 
In our main experiments, we also keep multiple candidate draft tokens at each drafting step, forming a draft tree that is subsequently verified in parallel by the full model using a tree-structured attention mask. To evaluate the contribution of this design to decoding efficiency, we ablate the tree-based verification mechanism by disabling tree attention and restricting drafting to a single candidate per step. We conduct this study on a GSM8K-finetuned \texttt{Qwen3-1.7B-Base} \methodname\ model and report results on GSM8K. As shown in Table~\ref{tab:tree_attn_ablation}, enabling tree-based verification consistently improves throughput, acceptance length, and acceptance rate across both fixed and dynamic drafting settings. These results demonstrate that expanding the candidate set and verifying branches in parallel substantially improves draft token acceptance, leading to higher overall decoding speedup.

\paragraph{Ablating Parameter Sharing.} To isolate the contribution of parameter sharing, we train an additional SEED control on GSM8K using \texttt{Qwen3-1.7B-Base} in which the drafter and verifier are fully parameter-disjoint, aiming to directly test whether enforced parameter sharing improves drafter-verifier alignment relative to an identically initialized but independently parameterized drafter. The two-layer drafter is initialized by copying the verifier's token embedding and tied output head, final normalization layer, and top two transformer layers. The drafter continues to consume the verifier-produced KV cache from layers 27 and 28, preserving the deep-feature conditioning mechanism. Both variants in this ablation use training block size $b=4$. All other training settings and objectives are unchanged. We evaluate both variants with a fixed draft length of 4 and without tree-based verification.

Table~\ref{tab:parameter_sharing} shows that removing parameter sharing substantially degrades both draft quality and task performance, despite the fact that this variant has more parameters and, in principle, greater expressive capacity. Although the independently parameterized drafter retains the same initialization and deep-feature conditioning mechanism, it still suffers from poorer draft token acceptance, resulting in reduced throughput. The degraded task performance further suggests that parameter sharing provides an important alignment benefit between the drafter and the verifier, making the joint optimization process more effective.

\begin{table}[htbp]
\centering
\caption{Parameter-sharing ablation on GSM8K with \texttt{Qwen3-1.7B-Base}. Removing parameter sharing substantially degrades both draft quality and task performance.}
\vspace{0.5em}
\label{tab:parameter_sharing}
\small
\setlength{\tabcolsep}{5pt}
\begin{tabular}{lcccc}
\toprule
 & \makecell{GSM8K\\Acc. (\%)} & \makecell{Throughput\\(tokens/s)} & \makecell{Draft Token\\Acc. Rate (\%)} & \makecell{Draft Token\\Acc. Length} \\
\midrule
\methodname\ & 57.5 & 74.3 & 73.0 & 2.9 \\
w/o Parameter Sharing & 44.1\drop{13.4} & 68.8\drop{5.5} & 68.5\drop{4.5} & 2.7\drop{0.2} \\
\bottomrule
\end{tabular}
\end{table}

\section{Additional Experimental Results}
\label{sec:add_exp_results}
\subsection{Why \methodname\ Improves Task Performance}
\label{sec:enc_hid_pred}

\begin{figure}[ht]
\centering

\begin{minipage}{0.48\textwidth}
\centering
\includegraphics[width=\linewidth]{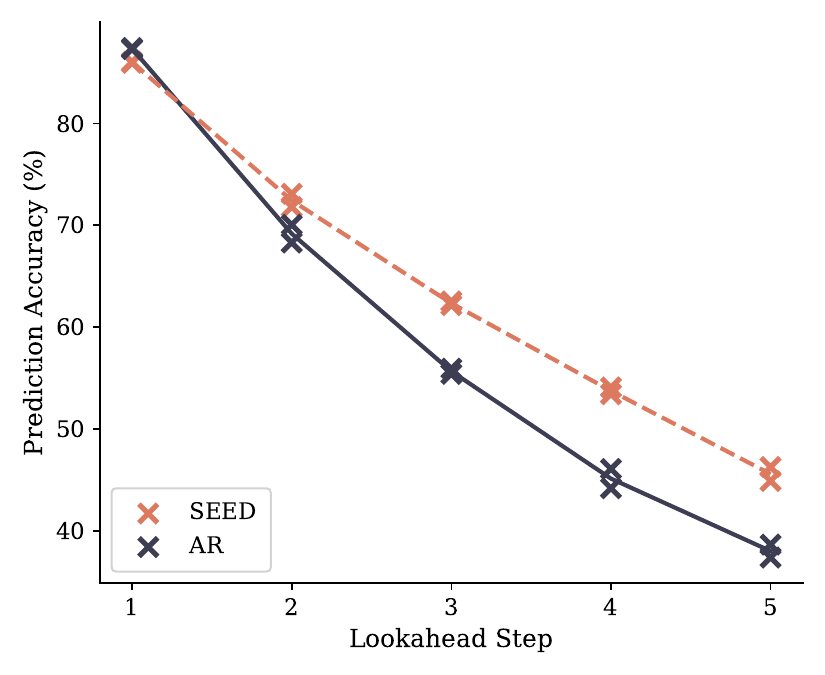}
\captionof{figure}{Training linear probes to predict future tokens from intermediate representations on Kodcode.}
\label{fig:enc_hid_pred_kodcode}
\end{minipage}
\hfill
\begin{minipage}{0.48\textwidth}
\centering
\includegraphics[width=\linewidth]{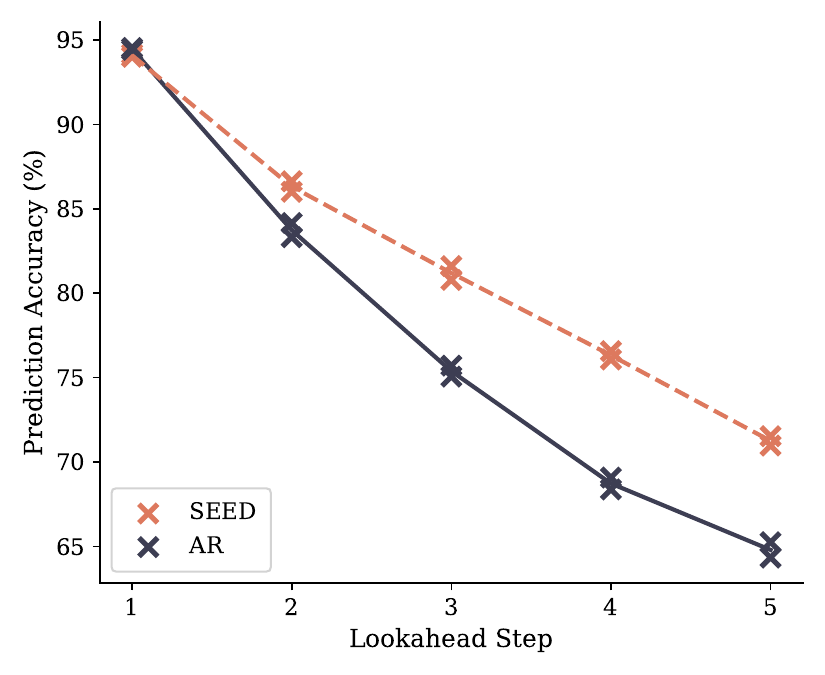}
\captionof{figure}{Training linear probes to predict future tokens from intermediate representations on ScienceQA.}
\label{fig:enc_hid_pred_scienceqa}
\end{minipage}

\end{figure}

Here we provide some further theoretical intuition as to why multi-token prediction (MTP) and \methodname\ achieve stronger task performance than standard autoregressive (AR) fine-tuning. We also include additional results of training linear probes to predict future tokens on two more datasets in Figures~\ref{fig:enc_hid_pred_kodcode} and~\ref{fig:enc_hid_pred_scienceqa}, using the intermediate representations of \texttt{Qwen3-1.7B-Base} AR and \methodname\ models.

In MTP-style training, the model learns to predict multiple future tokens from a single contextual representation. Concretely, given the hidden states of the final context token $h(t_0)$, the model learns to approximate $P(t_1, t_2, t_3 \mid h(t_0))$, forcing $h(t_0)$ to encode information that is predictive not only of the immediate next token but also of tokens further into the future, which encourages the model to implicitly plan ahead.

Similarly, during \methodname\ drafting, the decoder learns to model $P(t_1 \mid H^0_{t_0}, \kvcache{< t_0}{dec})$, $P(t_2 \mid H^0_{t_1}, H^0_{t_0}, \kvcache{< t_0}{dec})$, $P(t_3 \mid H^0_{t_2}, H^0_{t_1}, H^0_{t_0}, \kvcache{< t_0}{dec})$ and so on, where $H^0_{t_i}$ denotes the raw token embedding of token $t_i$. Importantly, these embeddings are fixed lookup vectors that cannot adapt to absorb predictive information during training. As a result, the gradients produced by the speculative loss $\mathcal{L}_{spec}$ primarily propagate through the cached contextual representations $\kvcache{< t_0}{dec}$, which originate from the encoder's hidden states. Consequently, the encoder is encouraged to compress information about upcoming tokens into its contextual representations so that the decoder can reliably predict future tokens during drafting.

For the linear probe experiments, we train both a \methodname\ model and an AR model with a block size of $4$ to examine the effect of planning ahead under a relatively short horizon. We then extract the representations from the final encoder layer $l'$ of the \methodname\ model and the corresponding layer $l'$ of its AR fine-tuned counterpart, run both models on the GSM8K test split, collect the hidden states produced at layer $l'$, and train five linear probes to independently predict future tokens at lookahead steps from $1$ to $5$. Specifically, we collect $20{,}000$ target-only token positions from $1{,}000$ samples in the test split and compute prediction accuracy against the ground-truth labels from the original datasets.

\subsection{Accelerating an Already Well-Tuned Model}
\label{sec:retrofit}

In our main experiments, \methodname\ is applied during task-specific fine-tuning, yielding both inference acceleration and improved task performance. A natural follow-up question is whether \methodname\ can be used purely to accelerate an already strong task model, without sacrificing its existing quality. To study this practical setting, we start from a \texttt{Qwen3-1.7B-Base} model that has already been fine-tuned with standard autoregressive (AR) training on GSM8K. We then perform an additional round of training to convert this AR model into a \methodname\ model. The training setup is identical to that described in Appendix~\ref{sec:training_details}, except that initialization comes from the task-tuned AR checkpoint rather than the untuned base model.

Table~\ref{tab:retrofit_ar} summarizes the results. Retrofitting \methodname\ onto an already fine-tuned AR model preserves task accuracy while achieving essentially the same decoding throughput as a \methodname\ model trained directly from the base model. The small accuracy gain is consistent with the MTP-like "planning ahead" effect induced by the speculative objective, as discussed in the main body.

These results are also aligned with our CKA analysis in Figure~\ref{fig:CKA_heatmap}, which shows that \methodname\ primarily reshapes internal model representations near the encoder–decoder interface while leaving earlier layers largely unchanged. This suggests that standard AR fine-tuning and \methodname\ training are structurally compatible: the latter can be placed on top of the former without disrupting previously learned task knowledge.

\begin{table}[h]
\centering
\caption{\methodname\ applied as a second-stage fine-tuning procedure on top of an AR-fine-tuned model. Dynamic drafting ($\tau=0.7$) and tree-based verification are used for \methodname\ models. Continued fine-tuning preserves accuracy on GSM8K while achieving comparable speedup.}
\label{tab:retrofit_ar}
\vspace{0.5em}
\resizebox{\textwidth}{!}{
\begin{tabular}{lcccc}
\toprule
\textbf{Model} & \textbf{Accuracy (\%)} & \textbf{Tput (tokens/s)} & \textbf{Acc. Len} & \textbf{Acc. Rate (\%)} \\
\midrule
AR (from base) & 56.3 & 37.9 & - & - \\
\methodname\ (continued from AR) & 56.5\up{0.2} & 90.6\up{52.7} & 3.91 & 86.82 \\
\methodname\ (from base) & 57.1 & 91.8 & 3.97 & 88.57 \\
\bottomrule
\end{tabular}
}
\end{table}

\section{Computational Cost Analysis}
\label{sec:computational_cost}
Here we provide a more comprehensive analysis of the theoretical computational cost of each method. For concreteness, we use \texttt{Qwen3-1.7B-Base} as an example. The model has $l=28$ transformer layers with $d_{\text{model}}=2048$, $d_{\text{ffn}}=6144$, 16 query heads, 8 KV heads, $d_{\text{head}}=128$, a vocabulary size of $|V|=151{,}936$, and tied input / output embeddings. Each attention block contains $W_Q\,(2048\times2048) + W_K\,(2048\times1024) + W_V\,(2048\times1024) + W_O\,(2048\times2048)$, for a total of $12.58$M parameters. The MLP block contributes $3 \times (2048\times6144)=37.75\text{M}$ parameters. Therefore, each transformer layer contains $P_{\text{layer}} = 50.33\text{M}$ parameters, and all 28 layers together contain $1.409$B parameters. The tied embedding / LM head contributes an additional $151936 \times 2048 = 311.2\text{M}$ parameters, resulting in a total of $P_{\text{full}} = 1.721\text{B}$ parameters. Throughout the following analysis, we normalize the computational cost of a full forward pass through the entire model as \emph{1 unit}, and express the cost of all compared methods relative to this baseline. Note that this parameter-based analysis is intended as a simple estimate of relative computational cost. It assumes that the amount of computation roughly scales with the number of parameters touched while ignoring operation-specific factors such as differences in FLOPs, memory access patterns, and hardware utilization across components.

\begin{table}[htbp]
\centering
\caption{Approximate drafting cost for a 1.7B model. Note that Apple MTP amortizes a full-model call over a draft block.}
\vspace{0.5em}
\label{tab:drafting_cost}
\small
\begin{tabular}{llrr}
\toprule
Method & Weights Touched per Draft Token & Weights & Cost (units) \\
\midrule
AR & 28 layers + head & 1721M & 1.000 \\
\methodname & 2 decoder layers + head & 411.9M & 0.239 \\
LayerSkip & 8 layers + head & 713.8M & 0.415 \\
Apple MTP & (28 layers + head) per block + sequential sampler head & $\geq1721$M & $\geq1.000$ \\
EAGLE-3 & 1 fused draft layer + reduced-vocab head & $\approx137$M & $\approx0.080$ \\
\bottomrule
\end{tabular}
\end{table}

As shown in Table~\ref{tab:drafting_cost}, \methodname's drafting step is $4.2\times$ cheaper than a full pass and $1.7\times$ cheaper than LayerSkip's early-exit pass, despite conditioning on strictly deeper representations.

Verification over $(d+1)$ tokens requires one full-model forward pass for all methods except LayerSkip. Since LayerSkip reuses the KV cache of the first 8 layers computed during drafting, verification only recomputes layers 9 - 28 together with the LM head. The corresponding cost is therefore $(20 \times 50.33 + 311.2) / 1721 = 0.766$ relative to a full forward pass.

Now consider a fixed draft length $d$ and per-token acceptance rate $r$. Each speculative decoding round incurs a computational cost of

$$
T_{\text{unit}} = d \cdot c_{\text{draft}} + c_{\text{verify}}
$$

where $c_{\text{draft}}$ and $c_{\text{verify}}$ denote the per-token drafting cost and the verification cost respectively. In return, each round commits an average of

$$
n = d \cdot r + 1
$$

tokens, consisting of the accepted draft tokens plus the bonus token generated during verification. Using the measured acceptance rates at $d=4$ (\methodname: 73.0\%, Apple MTP: 69.8\%, LayerSkip: 43.1\%) and taking one computational unit to correspond to a full autoregressive forward pass $(1/37.9~\text{s} = 26.4~\text{ms}$, based on the fine-tuned \texttt{Qwen3-1.7B-Base} achieving 37.9 tokens/s on GSM8K with a single A6000), we obtain the following throughput predictions by computing $1000 / T_{ms} \times n$:

\begin{table}[htbp]
\centering
\caption{Approximate throughput predictions for fixed draft length $d=4$ without tree-based verification. Apple MTP generates all four draft tokens in a single forward pass. However, its sampler applies the 311M-parameter LM head sequentially once per draft token, contributing an additional cost of $4 \times 311.2 / 1721 = 0.72$ units.}
\label{tab:cost_predictions}
\vspace{0.5em}
\small
\setlength{\tabcolsep}{4pt}
\begin{tabular}{lccccccc}
\toprule
Method & $c_{\mathrm{draft}}$ & $c_{\mathrm{verify}}$ & $T_{\mathrm{unit}}$ & \makecell{$T$\\(ms)} & $n$ & \makecell{Predicted\\(tokens/s)} & \makecell{Measured\\(tokens/s)} \\
\midrule
\methodname & 0.239 & 1.000 & 1.96 & 51.7 & 3.92 & \textbf{75.8} & \textbf{74.3} \\
LayerSkip & 0.415 & 0.766 & 2.43 & 64.1 & 2.72 & 42.4 & 47.0 \\
Apple MTP & $1.72/4$ & 1.000 & 2.72 & 71.8 & 3.79 & 52.8 & 44.6 \\
\bottomrule
\end{tabular}
\end{table}

\textbf{Despite its simplicity, this computational model closely predicts \methodname's measured throughput (w/ fixed draft length and w/o tree-based verification) and captures the relative performance trends across methods.} The remaining discrepancies particularly for LayerSkip and Apple MTP are expected because the model omits certain operation-specific costs and implementation overheads.

For completeness, EAGLE-3 achieves a substantially lower per-token drafting cost (0.08 units), owing to its lightweight single-layer drafter and reduced-vocabulary draft head. Therefore, \methodname's advantage does not primarily come from reducing the absolute drafting cost. Instead, it arises from the improved draft quality and higher acceptance rates.

The relative drafting cost of \methodname\ also becomes more favorable as the backbone model scales. For \texttt{Qwen3-4B-Base} (36 layers, $d_{\text{model}}=2560$, $P_{\text{layer}}=101$M, and a tied embedding / LM head of 389M parameters), the drafting cost is $(2\times101 + 389) / 4020 = 0.147$ units, compared with $0.239$ for \texttt{Qwen3-1.7B-Base}. This scaling behavior is consistent with the larger average speedup observed on the 4B model.

\section{Broader Impacts}
\label{sec:broader_impacts}
This work develops \methodname, a method for accelerating large language model (LLM) inference by improving self-speculative decoding. The primary positive societal impact is that faster inference can reduce the computational cost, latency, and energy consumption of deploying LLMs. More efficient inference may make advanced language models more accessible to researchers, developers, and users with limited computational resources, and may enable lower-cost applications in domains such as education, programming assistance, summarization, and scientific support.

At the same time, improving LLM inference efficiency can also amplify existing risks associated with language models. Lowering the cost and latency of generation could make it easier to produce large volumes of harmful or misleading content, including spam, phishing messages, disinformation, synthetic reviews, or other forms of automated manipulation. Faster inference may also facilitate misuse in surveillance, impersonation, or automated social engineering systems when combined with models capable of generating persuasive or personalized text.

Nevertheless, these risks are not unique to \methodname, but they are relevant because the method improves the efficiency of LLM generation. Future work on efficient inference should continue to evaluate not only speed and quality, but also whether acceleration changes the scale or detectability of potential misuse.

\end{document}